\documentclass[sigconf]{acmart}

\usepackage{booktabs} % For formal tables
\usepackage[ruled,vlined,linesnumbered]{algorithm2e}
\usepackage{multirow}
\usepackage{balance}
\usepackage{subcaption}
\usepackage{bbold}
\usepackage{enumitem}
\usepackage{xcolor, colortbl}
\usepackage{amsmath}

\newcommand{\jian}[1]{{\small\color{blue}{\bf Jian: #1}}}

\newcommand{\hide}[1]{}

\newcommand\blfootnote[1]{%
  \begingroup
  \renewcommand\thefootnote{}\footnote{#1}%
  \addtocounter{footnote}{-1}%
  \endgroup
}
\newcommand{\method}{\textsc{JuryGCN}}

\SetCommentSty{mycommentfont}

\setlist{leftmargin=10pt}
\DeclareMathOperator*{\argmin}{argmin}
\DeclareMathOperator*{\argmax}{argmax}

\definecolor{Gray}{gray}{0.75}
\newcolumntype{a}{>{\columncolor{Gray}}c}

\newtheorem{problem}{Problem}

\newtheorem{theorem}{Theorem}
\newtheorem{definition}{Definition}
\newtheorem{proposition}{Proposition}

\AtBeginDocument{%
  \providecommand\BibTeX{{%
    \normalfont B\kern-0.5em{\scshape i\kern-0.25em b}\kern-0.8em\TeX}}}
\copyrightyear{2022} 
\acmYear{2022} 
\setcopyright{acmcopyright}
\acmConference[KDD '22]{Proceedings of the 28th ACM SIGKDD Conference on Knowledge Discovery and Data Mining}{August 14--18, 2022}{Washington, DC, USA}
\acmBooktitle{Proceedings of the 28th ACM SIGKDD Conference on Knowledge Discovery and Data Mining (KDD '22), August 14--18, 2022, Washington, DC, USA}
\acmPrice{15.00}
\acmDOI{10.1145/3534678.3539286}
\acmISBN{978-1-4503-9385-0/22/08}
\begin{document}
% \fancyhead{}
% \mkclean{}
\title{\method: Quantifying Jackknife Uncertainty on Graph Convolutional Networks}

%% Author
\author{Jian Kang$^{*}$,~~~Qinghai Zhou$^{*}$, and Hanghang Tong}
\affiliation{
	\institution{
		University of Illinois at Urbana-Champaign, \{jiank2, qinghai2, htong\}@illinois.edu; \\
	}
	\country{}
}

\begin{abstract}
% \vspace{-1mm}
Graph Convolutional Network (GCN) has exhibited strong empirical performance in many real-world applications. %Despite its successful adoption of GCN in many application domains, 
The vast majority of existing works on GCN primarily focus on the accuracy while ignoring how confident or uncertain a GCN is with respect to its predictions. Despite being a cornerstone of trustworthy graph mining, uncertainty quantification on GCN has not been well studied and the scarce existing efforts either fail to provide deterministic quantification or have to change the training procedure of GCN by introducing additional parameters or architectures. In this paper, we propose the first frequentist-based approach named \method\ in quantifying the uncertainty of GCN, where the key idea is to quantify the uncertainty of a node as the width of confidence interval by a %\hh{Jackknife? check}\jian{jackknife is fine. most existing works does not put initial into uppercase.}
jackknife estimator. Moreover, we leverage the influence functions to estimate the change in GCN parameters without re-training to scale up the computation. %during when leaving out a node in the graph. 
% A distinctive advantage of the proposed \method\ lies in its capability to provide deterministic uncertainty quantification without modifying the GCN architecture or introduction of additional parameters \hh{this sentence is a bit too long and hard to understand}. 
The proposed \method\ is capable of quantifying uncertainty deterministically without modifying the GCN architecture or introducing additional parameters.
We perform extensive experimental evaluation on real-world datasets in the tasks of both active learning and semi-supervised node classification, which demonstrate the efficacy of the proposed method.

% \jian{found a recent ICML submission named `A General Framework for Quantifying Aleatoric and Epistemic Uncertainty in Graph Neural Networks' which include a frequentist model in experiments (written in abs), will keep an eye on arxiv}

\end{abstract}




\maketitle

%\hh{hide it for now}
\blfootnote{* Equal contribution.}

\vspace{-8mm}
\section{Introduction}~\label{sec:intro}
Graph Convolutional Network (GCN) has become a prevalent learning paradigm in many real-world applications, including %social network analysis~\cite{fan2019graph}, 
financial fraud detection~\cite{wang2019semi}, drug discovery~\cite{gaudelet2021utilizing} and traffic prediction~\cite{chen2019gated}. %Despite its satisfactory overall performance, 
To date, the vast majority of existing works do not take into account the uncertainty of a GCN regarding its prediction, which is alarming especially in high-stake scenarios. For example, in automated financial fraud detection, it is vital to let expert banker to take controls if a GCN-based detector is highly uncertain about its predictions in order to prevent wrong decisions on suspending banking account(s).

A well established study on uncertainty quantification of GCN could bring several crucial benefits. First, it is a cornerstone in trustworthy graph mining. Uncertainty quantification aims to understand to what extent the model is likely to be incorrect, and thus provides natural remedy to questions like \textit{how uncertain is a GCN in its own predictions?} Second, an accurate quantification of GCN uncertainty could potentially answer \textit{how to improve GCN predictions by leveraging its uncertainty} in many graph mining tasks. For example, in active learning on graphs, nodes with high uncertainty could be selected as the most valuable node to query the oracle; in node classification, the node uncertainty could help calibrate the confidence of GCN predictions, thereby improving the overall classification accuracy.

% \hh{is it true that "most of the existing work on GCNs focus on accuracy but ignore the confidence/uncertainty? if so, we may use this to open the paper. Then, we can talk why UQ is important (e.g., (1) cornerstone of trustworthy GM; (2) some other GM methods such as active learning depend on uncertainty}\jian{i think it is true. i will reformulate the first few paragraphs.}
% Graphs naturally arise in many real-world applications, including social network analysis~\cite{fan2019graph}, financial fraud detection~\cite{zhang2017hidden}, drug discovery~\cite{gaudelet2021utilizing} and traffic prediction~\cite{chen2019gated}. The ubiquity of graph structured data makes it important to gain deep insights and understandings from graphs through sophisticated graph mining models. Among them, Graph Convolutional Network (GCN)~\cite{kipf2017semi} is one of the most representative and widely used ones. 

% Despite the satisfactory overall performance of GCN in learning node representations, it is also desired to understand how certain a GCN is likely to be correct, especially in high-stake scenarios. For example, in automated financial fraud detection where a GCN is applied to identify suspicious banking activities (e.g., money laundering), controls by expert banker may be required if the GCN is uncertain about its predictions. In particular, it remains opaque on questions like \textit{how certain/uncertain is GCN to be correct? How can we improve GCN predictions by leveraging its uncertainty?}

Important as it could be, very few studies on uncertainty quantification of GCN exist, which mainly focuses on two different directions: Bayesian-based approaches and deterministic quantification-based approaches. Regarding Bayesian-based approaches~\cite{hasanzadeh2020bayesian, zhang2019bayesian}, they either drops edge(s) with certain sampling strategies or leverages random graph model (e.g., stochastic block model) to assign edge probabilities for training. However, these models fall short in explicitly quantifying the uncertainty on model predictions. Another type of methods, i.e., deterministic quantification-based approaches~\cite{liu2020uncertainty, zhao2020uncertainty, stadler2021graph}, directly quantifies uncertainty by parameterizing a Dirichlet distribution as prior to estimate the posterior distribution under a Bayesian framework. Nevertheless, it changes the training procedures of a graph neural network by introducing additional parameters (e.g., parameters for Dirichlet distribution) or additional architectures (e.g., teacher network) in order to precisely estimate the uncertainty.

To address the aforementioned limitations, we provide the first study on frequentist-based analysis of the GCN uncertainty, which we term as the \method\ problem. Building upon the general principle of jackknife (leave-one-out) resampling~\cite{miller1974jackknife}, the jackknife uncertainty of a node is defined as the width of confidence interval constructed by a jackknife estimator %\hh{jackknife or jackknife resampling or jackknife estimator? check and pick one and use it consistently}\jian{jackknife resampling is a technique which is achieved by a jackknife estimator. i think it might be more appropriate to say `the jackknife estimator constructs the confidence interval' and `jackknife resampling is a general principle of many jackknife estimators'.}
when leaving the corresponding node out. In order to estimate the GCN parameters without exhaustively re-training GCN, we leverage influence functions~\cite{koh2017understanding} to quantify the change in GCN parameters by infinitesimally upweighting the loss of a training node. Compared with existing works, our method brings several advantages. First, our method provides deterministic uncertainty quantification, which is not available in Bayesian-based approaches~\cite{hasanzadeh2020bayesian, zhang2019bayesian}. Second, different from existing works on deterministic uncertainty quantification~\cite{liu2020uncertainty, zhao2020uncertainty, stadler2021graph}, our method does not introduce any additional parameters or components in the GCN architecture. Third, our method can provide \textit{post-hoc} uncertainty quantification. As long as the input graph and a GCN are provided, our method can \textit{always} quantify node uncertainty without any epoch(s) of model training.

The major contributions of this paper are summarized as follows.
\vspace{-4mm}
\begin{itemize}[
    align=left,
    leftmargin=2em,
    itemindent=0pt,
    labelsep=0pt,
    labelwidth=1em,
]
\vspace{-1mm}
    \item \textbf{Problem definition.} To our best knowledge, we provide the first frequentist-based analysis of GCN uncertainty and formally define the \method\ problem.
    \item \textbf{Algorithm and analysis.} We propose \method\ to quantify jackknife uncertainty on GCN. The key idea is to leverage a jackknife estimator to construct a leave-one-out predictive confidence interval for each node, where the leave-one-out predictions are estimated using the influence functions with respect to model parameters. 
    \item \textbf{Experimental evaluations.} We demonstrate the effectiveness of \method~ through extensive experiments on real-world graphs in active learning on node classification and semi-supervised node classification.
\end{itemize}

% The rest of this paper is organized as follows. We formally define the problem of \method\ in Section~\ref{sec:problem}. Section~\ref{sec:measure} presents the definition of jackknife uncertainty and our mathematical analysis on influence function computation. Our proposed \method\ is shown in Section~\ref{sec:algorithm}. Section~\ref{sec:experiment} includes the experimental settings and evaluation results. After reviewing related work in Section~\ref{sec:related}, we conclude the paper in Section~\ref{sec:conclusion}. \jian{can remove this paragraph if significantly out of page}
\vspace{-4mm}
\section{Problem Definition}~\label{sec:problem}
%\hh{i think it is fine if we call this section just 'problem definition'}
% In this section, we first present the key symbols of this paper in Table~\ref{tab:symbols}. After that, we introduce preliminary knowledge on the Graph Convolutional Network (GCN), predictive uncertainty and jackknife resampling. Finally, we formally define the problem of jackknife uncertainty quantification on GCN (\method).
In this section, we first introduce preliminary knowledge on the Graph Convolutional Network (GCN), predictive uncertainty and jackknife resampling. Then, we formally define the problem of jackknife uncertainty quantification on GCN (\method).

Unless otherwise specified, we use bold upper-case letters for matrices (e.g., $\mathbf{A}$), bold lower-case letters for vectors (e.g., $\mathbf{x}$), calligraphic letters for sets (e.g., $\mathcal{G}$) and fraktur font for high-dimensional tensors ($\mathfrak{H}$). We use supercript $^T$ for matrix transpose and superscript $^{-1}$ for matrix inversion, i.e., $\mathbf{A}^T$ and $\mathbf{A}^{-1}$ are the transpose and inverse of $\mathbf{A}$, respectively. We use conventions similar to PyTorch in Python for indexing. For example, $\mathbf{A}[i,j]$ represents the entry of $\mathbf{A}$ at the $i$-th row and $j$-th column; $\mathbf{A}[i,:]$ and $\mathbf{A}[:,j]$ demonstrate the $i$-th row and $j$-th column of $\mathbf{A}$, respectively.
\vspace{-4mm}

% \begin{table}[htp]
%     \centering
%     \caption{Symbols and Notations.\qh{maybe we can hide the notation table since most notations are clearly introduced in the main text.}}
%     % \vspace{-4mm}
%     \scalebox{0.9}{
%     \begin{tabular}{c|c}
%         \hline
%         \textbf{Symbol} & \textbf{Description} \\
%         \hline
%         $\mathcal{G}$ & a graph \\
%         $\mathcal{V}$ & a set of nodes \\
%         $\mathcal{Y}$ & a set of node labels \\
%         $\mathbf{A}$, $\mathbf{X}$ & adjacency, node feature matrix \\
%         % $\mathbf{X}$ & node feature matrix \\
%         \hline 
%         $L$ & number of hidden layers \\
%         $\Theta$ & model parameters \\
%         $\textit{GCN}()$ & an $L$-layer GCN \\
%         $\mathbf{E}^{(l)}$ & node representations at the $l$-th layer \\
%         $\mathbf{W}^{(l)}$ & weight matrix at the $l$-th layer \\
%         $\sigma()$ & activation function \\
%         \hline 
%         $\alpha$ & coverage parameter \\
%         $Q_{\alpha}()$ & the $\alpha$ quantile\\
%         $\mathbb{I}_{\Theta}()$ & influence of a node with respect to $\Theta$ \\
%         $\mathbb{U}_\Theta()$ & uncertainty of a node \\
%         \hline
%     \end{tabular}}
%     \label{tab:symbols}
% \end{table}
% \vspace{-6mm}
\subsection{Preliminaries}
\vspace{-1mm}
\noindent \textbf{1 -- Graph Convolutional Network (GCN)} 
% is one of the most fundamental graph neural networks. It 
% learns node representations by a collection of graph convolution operations. Graph convolution usually consists of two key steps. Each node (1) aggregates its own representation with the representations of its neighbors and then (2) transforms its aggregated representations by a fully connected layer. 
Let $\mathcal{G} = \{\mathcal{V}, \mathbf{A}, \mathbf{X}\}$ denote a graph whose node set is $\mathcal{V}$, adjacency matrix is $\mathbf{A}$ and node feature matrix is $\mathbf{X}$. For the $l$-th hidden layer in an $L$-layer GCN, we assume $\mathbf{E}^{(l)}$ is the output node embeddings (where $\mathbf{E}^{(0)} = \mathbf{X}$) and $\mathbf{W}^{(l)}$ is the weight matrix. 
% For notation clarity and consistency, we let $\mathbf{E}^{(0)} = \mathbf{X}$. 
Mathematically, the graph convolution at the $l$-th hidden layer can be represented by $\mathbf{E}^{(l)} = \sigma(\mathbf{\hat A} \mathbf{E}^{(l-1)} \mathbf{W}^{(l)})$ where $\sigma$ is the activation and $\mathbf{\hat A} = \mathbf{\tilde D}^{-\frac{1}{2}} (\mathbf{A} + \mathbf{I}) \mathbf{\tilde D}^{-\frac{1}{2}}$ is the renormalized graph Laplacian with $\mathbf{\tilde D}$ being the degree matrix of $(\mathbf{A} + \mathbf{I})$.

\noindent \textbf{2 -- Uncertainty quantification} is one of the cornerstones 
% for trustworthy data mining 
in safe-critical applications. It provides accurate quantification on how confident a mining model is towards its predictions. In general, uncertainty can be divided into two types: \textit{aleatoric} uncertainty and \textit{epistemic} uncertainty~\cite{abdar2021review}. Aleatoric uncertainty (or data uncertainty) refers to the variability in mining results due to the inherent randomness in input data, which is irreducible due to complexity of input data (e.g., noise); whereas epistemic uncertainty (or model uncertainty) measures how well the mining model fits the training data due to the lack of knowledge on the optimal model parameters, which is reducible by increasing the size of training data. 

\noindent \textbf{3 -- Jackknife resampling} is a classic method to estimate the bias and variance of a population~\cite{tukey1958bias}. It often relies on a jackknife estimator which is built by leaving out an observation from the entire population (i.e., leave-one-out) and evaluating the error of the model re-trained on the held-out population. Suppose we have (1) a set of $n$ data points $\mathcal{D}=\{(\mathbf{x}_i, y_i) | i=1,\ldots,n\}$\hide{ with $\mathbf{x}_i$ and $y_i$ as the feature and label of the $i$-th data point}, (2) a test point $(\mathbf{x}_{\textrm{test}}, y_{\textrm{test}})$, (3) a mining model $f_\theta()$ parameterized by $\theta$ (e.g., a neural network) where $f_\theta(\mathbf{x})$ is the prediction of input feature $\mathbf{x}$ and (4) a target coverage level $(1-\alpha)$ such that the label $y$ is covered by the predictive confidence interval with probability $(1-\alpha)$. Mathematically, the
% the upper bound $\mathbb{C}^+(\mathbf{x}_{\textrm{test}})$ and lower bound $\mathbb{C}^-(\mathbf{x}_{\textrm{test}})$ of the predictive 
confidence interval constructed by the naive jackknife~\cite{efron1992jackknife} is 
upper bounded by $\mathbb{C}^+(\mathbf{x}_{\textrm{test}}) = Q_{1-\alpha}(\mathcal{R}^{+})$ and lower bounded by $\mathbb{C}^-(\mathbf{x}_{\textrm{test}}) = Q_{\alpha}(\mathcal{R}^{-})$, 
% \begin{equation}\label{eq:naive_jackknife}
%     \mathbb{C}^+(\mathbf{x}_{\textrm{test}}) = Q_{1-\alpha}(\mathcal{R}^{+}) \qquad \mathbb{C}^-(\mathbf{x}_{\textrm{test}}) = Q_{\alpha}(\mathcal{R}^{-})
% \end{equation}
where $Q_{\alpha}$ finds the $\alpha$ quantile of a set and $\mathcal{R}^{\gamma} = \{f_{\theta}(\mathbf{x}_{\textrm{test}}) + \gamma\cdot\lvert y_{\textrm{test}} - f_{\theta_{-i}}(\mathbf{x}_{\textrm{test}}) \rvert | i=1, \ldots, n\}$ for $\gamma\in\{-, +\}$ and $\lvert y_{\textrm{test}} - f_{\theta_{-i}}(\mathbf{x}_{\textrm{test}}) \rvert$ is the error residual of the re-trained model on the dataset $\mathcal{D}\setminus \{(\mathbf{x}_i, y_i)\}$ (i.e., parameterized by $\theta_{-i}$).\footnote{We use $\gamma$ to represent the symbol before the leave-one-out error, i.e., $f_{\theta}(\mathbf{x}_{\textrm{test}}) -\lvert y_{\textrm{test}} - f_{\theta_{-i}}(\mathbf{x}_{\textrm{test}}) \rvert$ when $\gamma=-$, or $f_{\theta}(\mathbf{x}_{\textrm{test}}) + \lvert y_{\textrm{test}} - f_{\theta_{-i}}(\mathbf{x}_{\textrm{test}}) \rvert$ otherwise.} Hence, $\mathcal{R}^+$ and $\mathcal{R}^-$ represent the sets of upper and lower uncertainty bound on the original model prediction (i.e., $f_\theta(\mathbf{x}_{\mathrm{test}})$). Furthermore, jackknife+~\cite{barber2021predictive} constructs the predictive confidence interval for exchangeable data as 
\vspace{-1mm}
\begin{equation}\label{eq:jackknife+}
    \mathbb{C}^+(\mathbf{x}_{\textrm{test}}) = Q_{1-\alpha}(\mathcal{P}^{+}) \qquad \mathbb{C}^-(\mathbf{x}_{\textrm{test}}) = Q_{\alpha}(\mathcal{P}^{-})
\vspace{-1mm}
\end{equation}
where $\mathcal{P}^{\gamma}$ for $\gamma\in\{-, +\}$ is defined as $\mathcal{P}^{\gamma} = \{f_{\theta_{-i}}(\mathbf{x}_{\textrm{test}}) + \gamma\cdot \lvert y_i - f_{\theta_{-i}}(\mathbf{x}_i)\rvert | i=1, \ldots, n \}$. Similarly, %after having obtained the error residual $\lvert y_{\textrm{test}} - f_{\theta_{-i}}(\mathbf{x}_{\textrm{test}})\rvert$ of the re-trained model (i.e., $f_{\theta_{-i}}(\cdot)$), 
$\mathcal{P}^-$ and $\mathcal{P}^+$ represent the sets of the lower and upper uncertainty bound of the leave-one-out prediction $f_{\theta_{-i}}(\mathbf{x}_{\textrm{test}})$, respectively. With the assumption on data exchangeability, it yields a $(1-2\alpha)$ coverage rate theoretically.
\vspace{-3mm}
\subsection{Problem Definition}
\vspace{-1mm}
Existing works on deterministic uncertainty quantification on a graph neural network (GNN) mainly rely on changing the training procedures of a vanilla GNN (i.e., graph neural network without consideration of uncertainty)~\cite{zhao2020uncertainty, stadler2021graph}. As such, given a well-trained GNN, it requires epoch(s) of re-training to quantify its uncertainty. Nevertheless, it would cost a lot of computational resources to re-train it, especially when the model has already been deployed in an operational environment. Additionally, it remains a challenging problem to further comprehend the predictive results of GNN from the perspective of uncertainty, and to answer the following question: \textit{to what extent the GNN model is confident of the current prediction?} Therefore, it is essential to investigate the uncertainty quantification in a post-hoc manner, i.e., quantifying uncertainty without further (re-)training on the model.

Regarding post-hoc uncertainty quantification for IID (i.e., non-graph) data, Alaa and van der Schaar~\cite{alaa2020discriminative} propose a frequentist-based method inspired by jackknife resampling. It uses high-order influence functions to quantify the impact of a data point on the underlying neural network. Given that the parameters of a neural network are derived by learning with data points, high-order influence functions are capable of understanding how much a data point will affect the model parameters. Then, the change in model parameters can be used to infer the uncertainty of the corresponding data point on the neural network by a jackknife estimator~\cite{barber2021predictive}. Under mild assumption (such as the algorithmic stability assumption and the IID/exchangebility of data), the naive jackknife estimator~\cite{efron1992jackknife} and its variants~\cite{barber2021predictive} bear strong theoretical guarantee in terms of the coverage such that the confidence interval will cover the true model parameters with a high probability. %\hh{might be a goode idea to add a few sentences on the theoretic properties of js, e.g., (we can use a similar narrative as in the proposal), e.g., under some mild assumption (such as xxx), JS and its variants bear strong theoretic properties in terms of xxx and xxx ~\cite{}. }

Building upon the jackknife resampling~\cite{miller1974jackknife} and the general principle outlined in~\cite{alaa2020discriminative}, we seek to bridge the gap between frequentist-based uncertainty quantification and graph neural networks. To be specific, given an input graph and a GCN, we aim to estimate the uncertainty of a node as the impact of leaving out its loss when computing the overall loss. Formally, we define the problem of jackknife uncertainty quantification on GCN, which is referred to as \method\ problem.
\vspace{-1mm}
\begin{problem}
	\method: \underline{J}ackknife \underline{U}nce\underline{r}taint\underline{y} Quantification on \underline{G}raph \underline{C}onvolutional \underline{N}etwork
\end{problem}
\vspace{-2mm}
\textbf{Given:} (1) An undirected graph $\mathcal{G}=\{\mathcal{V}, \mathbf{A}, \mathbf{X}\}$; (2) an $L$-layer GCN with the set of weights $\Theta$;  (3) a task-specific loss function $R(\mathcal{G}, \mathcal{Y}, \Theta)$ where $\mathcal{Y}$ is the set of node labels.

\textbf{Find:} An uncertainty score $\mathbb{U}_{\Theta}(u)$ for any node $u$ in graph $\mathcal{G}$ with respect to the GCN parameters $\Theta$ and the task-specific loss function $R(\mathcal{G}, \mathcal{Y}, \Theta)$.

% \hh{overall impression until here: the logic of the paper is quite clear now. the writing could be further smoothed (e.g., some sentences are way too long; some sentences read a bit awkward with many negations (existing works do not take into account how uncertain a gcn is likely to be incorrect))}
\vspace{-3mm}
\section{\method: Measure and Computation}~\label{sec:measure}
% In this section, we present an end-to-end algorithm \hh{this is in sec 4 now? check}\jian{yes, its in section 4, i will change the text accordingly}, namely \method, to quantify the jackknife uncertainty on GCN. We start by discussing the general strategy of quantifying jackknife uncertainty with influence functions, and then present analysis on computation of jackknife uncertainty in GCN. %Based on that, we propose \method\ algorithm for jackknife uncertainty quantification.%\hh{this is HUGE section with about 5 pages. let's consider to split it into two, e.g., sec 3: JuryGCN: Jackknife Uncertainty and Influence Function (or Measure and Computation); sec 4: JuryGCN: Algorithm and Applications (current 3.3)}
In this section, we start by discussing the general strategy of quantifying jackknife uncertainty with influence functions, and formally define the node jackknife uncertainty. After that, we present mathematical analysis on the influence function computation for GCN.
\vspace{-6mm}
\subsection{Jackknife Uncertainty of GCN}
\vspace{-1mm}
In this paper, we consider an $L$-layer GCN with ReLU activation for node-level tasks (e.g., node classification). We further assume that the nodes of the input graph $\mathcal{G}$ are exchangeable data.\footnote{A sequence of random variables is exchangeable if and only if the joint distribution of the random variables remains unchanged regardless of their ordering in the sequence~\cite{vovk2005algorithmic}. This assumption is commonly used in random graph models, e.g., Erd{\H{o}}s-R{\'e}nyi model~\cite{erdHos1960evolution}, stochastic block model~\cite{holland1983stochastic}.}

We first observe that the loss function of many node-level tasks can often be decomposed into a set of node-specific subproblems. Mathematically, it can be written as %the following optimization problem.
\begin{equation}\label{eq:optimization}
\displaystyle    \Theta^* = \argmin_{\Theta} R(\mathcal{G}, \mathcal{Y}_{\textrm{train}}, \Theta) = \argmin_{\Theta}\frac{1}{|\mathcal{V}_{\textrm{train}}|} \sum_{v\in\mathcal{V}_{\textrm{train}}}r(v, \mathbf{y}_v, \Theta)
\end{equation}
where $\mathcal{V}_{\textrm{train}} \subseteq \mathcal{V}$ is the set of training nodes, $|\mathcal{V}_{\textrm{train}}|$ is the number of training nodes, $\mathcal{Y}_{\textrm{train}} = \{\mathbf{y}_v | v\in\mathcal{V}_{\textrm{train}}\}$ is the set of ground-truth training labels and $\mathbf{y}_v$ is the label of node $v$. In this case, the overall loss function $R(\mathcal{G}, \mathcal{Y}_{\textrm{train}}, \Theta)$ is decomposed into several subproblems, each of which minimizes the node-specific loss function $r(v, \mathbf{y}_v, \Theta)$ for a node $v$. An example is the cross entropy with node-specific loss as $r(v, \mathbf{y}_v, \Theta) = -\sum_{i=1}^{c} \mathbf{y}_v[c] \log\big(\textit{GCN}(v, \Theta)[c]\big)$, 
where $c$ is the number of classes, $\textit{GCN}(v, \Theta)$ is the vector of predicted probabilities for each class using the GCN with parameters $\Theta$. If we upweight the importance of optimizing the loss of a node $i$ with some small constant $\epsilon$, we have the following loss function.
\begin{equation}\label{eq:perturbed_optimization}
    \Theta^*_{\epsilon, i} = 
    \argmin_{\Theta} 
    \epsilon r(i, \mathbf{y}_i, \Theta) 
    + 
    \frac{1}{|\mathcal{V}_{\textrm{train}}|} 
    \sum_{v \in \mathcal{V}_{\textrm{train}}} 
        r(v, \mathbf{y}_v, \Theta) 
\vspace{-2mm}
\end{equation}
Influence function is a powerful approach to evaluate the dependence of the estimator on the value of the data examples~\cite{koh2017understanding,zhou2021attent}. In order to obtain $\Theta^*_{\epsilon, i}$ without re-training the GCN, we leverage the influence function~\cite{koh2017understanding}, which is essentially the Taylor expansion over the model parameters. 
\begin{equation}\label{eq:taylor}
    \Theta^*_{\epsilon, i} \approx \Theta^* + \epsilon \mathbb{I}_{\Theta^*}(i)
\vspace{-4mm}
\end{equation}
where $\mathbb{I}_{\Theta^*}(i)=\frac{d \Theta^*_{\epsilon, i}}{d \epsilon}|_{\epsilon=0}$ is the influence function with respect to node $i$. The influence function $\mathbb{I}_{\Theta^*}(i)$ can be further computed using the classical result in \cite{cook1982residuals} as
\begin{equation}\label{eq:node_influence}
    \mathbb{I}_{\Theta^*}(i) = \mathbf{H}_{\Theta^*}^{-1}\nabla_{\Theta} r(i, \mathbf{y}_i, \Theta^*)
\end{equation}
where $\mathbf{H}_{\Theta^*} = \frac{1}{|\mathcal{V}_{\textrm{train}}|} \nabla^2_\Theta R(\mathcal{G}, \mathcal{Y}_{\textrm{train}}, \Theta^*)$ is the Hessian matrix with respect to model parameters $\Theta^*$. For a training node $i$, by setting $\epsilon = - \frac{1}{|\mathcal{V}_{\textrm{train}}|}$, Eq.~\eqref{eq:taylor} efficiently estimates the leave-one-out (LOO) parameters $\Theta^*_{\epsilon, i}$ if leaving out the loss of node $i$. After that, by simply switching the original model parameters to $\Theta^*_{\epsilon, i}$, we estimate the leave-one-out error $\textit{err}_i$ of node $i$ as follows.
\begin{equation}\label{eq:loo_error}
    \textit{err}_i = \|\mathbf{y}_i - \textit{GCN}(i, \Theta^*_{\epsilon, i})\|_2
\end{equation}
where $\textit{GCN}(u, \Theta^*_{\epsilon, i})$ represents the output of node $u$ using the GCN with leave-one-out parameters $\Theta^*_{\epsilon, i}$.

With Eq.~\eqref{eq:loo_error}, we use jackknife+~\cite{barber2021predictive}, which requires the data to be exchangeable instead of IID, to construct the confidence interval of node $u$. Mathematically, the lower bound $\mathbb{C}_{\Theta}^{-}(u)$ and upper bound $\mathbb{C}_{\Theta}^{+}(u)$ of the predictive confidence interval of node $u$ are
\begin{equation}\label{eq:ci}
\begin{array}{c}
    \mathbb{C}_{\Theta^*}^{-}(u) = Q_{\alpha}(\{\|\textit{GCN}(u, \Theta^*_{\epsilon, i})\|_2 - \textit{err}_i | \forall i\in\mathcal{V}_{\textrm{train}}\setminus\{u\}\}) \\
    \mathbb{C}_{\Theta^*}^{+}(u) = Q_{1-\alpha}(\{\|\textit{GCN}(u, \Theta^*_{\epsilon, i})\|_2 + \textit{err}_i | \forall i\in\mathcal{V}_{\textrm{train}}\setminus\{u\}\})
\end{array}
\end{equation}
where $Q_{\alpha}$ and $Q_{1-\alpha}$ are the $\alpha$ and $(1-\alpha)$ quantile of a set. Since a wide confidence interval of node $u$ means that the model is less confident with respect to node $u$, it implies that node $u$ has high uncertainty. Following this intuition, the uncertainty of node $u$ can be naturally quantified by the width of the corresponding confidence interval (Eq.~\eqref{eq:ci}). Since the uncertainty is quantified using the confidence interval constructed by a jackknife estimator, we term it as \textit{jackknife uncertainty} which is formally defined in Definition~\ref{defn:jackknife_uncertainty}.
\vspace{-2mm}
\begin{definition}\label{defn:jackknife_uncertainty}
    (Node Jackknife Uncertainty) Given an input graph $\mathcal{G}$ with node set $\mathcal{V}$, a set of training nodes $\mathcal{V}_{\textrm{train}} \subseteq \mathcal{V}$ and an $L$-layer GCN with parameters $\Theta$, $\forall i\in\mathcal{V}_{\textrm{train}}$ and $\forall u\in\mathcal{V}$, we assume (1) the nodes are exchangeable, and denote that (2) the LOO parameters are $\Theta_{\epsilon, i}$, (3) the error is defined as Eq.~\eqref{eq:loo_error} and (4) the lower bound $\mathbb{C}_{\Theta}^{-}(u)$ and the upper bound $\mathbb{C}_{\Theta}^{+}(u)$ of predictive confidence interval are defined as Eq.~\eqref{eq:ci}, the jackknife uncertainty of node $u$ is
    % Assuming the exchangeability of nodes, suppose we have an input graph $\mathcal{G}$ with node set $\mathcal{V}$, a set of training node $\mathcal{V}_{\textrm{train}} \subseteq \mathcal{V}$ and an $L$-layer GCN with parameters $\Theta$. 
    % $\forall i\in\mathcal{V}_{\textrm{train}}$. Supposing that the LOO parameters are $\Theta_{\epsilon, i}$, the error is defined as Eq.~\eqref{eq:loo_error}, and the lower bound $\mathbb{C}_{\Theta}^{-}(u)$ and upper bound $\mathbb{C}_{\Theta}^{+}(u)$ of predictive confidence interval are defined as Eq.~\ref{eq:ci} %as  \hh{1.eq10 is exactly the same as eq9? 2. for the second line, it should be $Q_{1-\alpha}$?}\jian{yes, i will change the text and refer to Eq.10}
    % \begin{equation}
    % \begin{array}{c}
    %     \mathbb{C}_{\Theta}^{-}(u) = Q_{\alpha}(\{\|\textit{GCN}(u, \Theta_{\epsilon, i})\|_2 - \textit{err}_i | i\in\mathcal{V}_{\textrm{train}}\}) \\
    %     \mathbb{C}_{\Theta}^{+}(u) = Q_{1-\alpha}(\{\|\textit{GCN}(u, \Theta_{\epsilon, i})\|_2 + \textit{err}_i | i\in\mathcal{V}_{\textrm{train}}\}) \\
    % \end{array}
    % \end{equation}
    % where $Q_{\alpha}$ and $Q_{1-\alpha}$ are the $\alpha$ and $(1-\alpha)$ quantile of a set. The
    % the jackknife uncertainty of any node $u\in\mathcal{V}$ is 
    \begin{equation}\label{eq:jackknife_uncertainty}
        \mathbb{U}_\Theta(u) = \mathbb{C}_{\Theta}^{+}(u) - \mathbb{C}_{\Theta}^{-}(u)
    \end{equation}
\end{definition}

We note that Alaa and van Der Schaar~\cite{alaa2020discriminative} leverage high-order influence functions to quantify the jackknife uncertainty for IID data. Though Eq.~\eqref{eq:taylor} shares the same form as in \cite{alaa2020discriminative} when the order is up to 1, our work bears three subtle differences. First, \cite{alaa2020discriminative} views the model parameters as statistical functionals of data distribution and exploits von Mises expansion over the data distribution to estimate the LOO parameters,\footnote{A statistical functional is a map that maps a distribution to a real number.} which is fundamentally different from our Taylor expansion-based estimation. Specifically, von Mises expansion requires that the perturbed data distribution should be in a convex set of the original data distribution and all possible empirical distributions~\cite{fernholz2012mises}. Since the node distribution of a graph is often unknown, the basic assumption of von Mises expansion might not hold on graph data. However, our definition relies on the Taylor expansion over model parameters which are often drawn independently from Gaussian distribution(s). Thus, our method is able to generalize on graphs. Second, \cite{alaa2020discriminative} works for regression or binary classification tasks by default, whereas we target more general learning settings on graphs (e.g., multi-class node classification). Third, jackknife uncertainty is always able to quantify aleatoric uncertainty and epistemic uncertainty simultaneously on IID data. Nevertheless, as shown in Proposition~\ref{prop:both_types_on_GCN}, it requires additional assumption to quantify both types of uncertainty on GCN simultaneously for a node $u$. 
\vspace{-2mm}
\begin{proposition}\label{prop:both_types_on_GCN}
(Necessary condition of aleatoric and epistemic uncertainty quantification on GCN) Given an input graph $\mathcal{G}$ whose node set is $\mathcal{V}$, a node $u\in\mathcal{V}$, a set of training nodes $\mathcal{V}_{\textrm{train}}$ and an $L$-layer GCN, jackknife uncertainty quantifies the aleatoric uncertainty and the epistemic uncertainty as long as $u$ is outside the $L$-hop neighborhood of an arbitrary training node $v\in\mathcal{V}_{\textrm{train}}\setminus\{u\}$. %that is not node $u$.
\end{proposition}
\vspace{-4mm}
\begin{proof}
See Appendix.
\end{proof}
\vspace{-2mm}
\noindent \textit{Remark.} For GCN, jackknife uncertainty cannot always measure the aleatoric uncertainty and epistemic uncertainty simultaneously. In fact, if a node $u$ is one of the neighbors within $L$ hops with respect to any training node $v\in\mathcal{V}_{\textrm{train}}\setminus\{u\}$, jackknife uncertainty only quantifies the epistemic uncertainty due to lack of knowledge on the loss of node $u$. In this case, jackknife uncertainty cannot quantify aleatoric uncertainty because leaving out the loss of node $u$ does not necessarily remove node $u$ in the graph. More specifically, the aleatoric uncertainty of node $u$ can still be transferred to its neighbors through neighborhood aggregation in graph convolution.

\vspace{-3mm}

\subsection{The Influence Functions of GCN}
\vspace{-1mm}
In order to quantify jackknife uncertainty (Eq.~\eqref{eq:jackknife_uncertainty}), we need to compute the influence functions to estimate the leave-one-out parameters by Eq.~\eqref{eq:taylor}. Given a GCN with $\Theta$ being the set of model parameters, to compute the influence functions of node $i$ (Eq.~\eqref{eq:node_influence}), we need to compute two key terms, including (1) first-order derivative $\nabla_{\Theta}r(i, \mathbf{y}_i, \Theta)$ and (2) second-order derivative $\mathbf{H}_{\Theta}$. 
We first give the following proposition for the computation of first-order derivative in Proposition~\ref{lm:first_order}.\footnote{The method to compute the first-order derivative was first proposed in \cite{kang2022rawlsgcn} for a different purpose, i.e., ensuring degree-related fairness in GCN.} Based on that, we present the main results for computing the second-order derivative in Theorem~\ref{lm:second_order}. Finally, we show details of influence function computation in Algorithm~\ref{alg:hvp}. %\hh{let's have a small 'roadmap' paragraph here -- this might make the relationship between lemma 1 and lemma 2 more clear (e.g., the implicit message is that, yes, lemma 1 is not our main contribution, but we need it. but we have a much more substantial new contribution in lemma 2). you migh also consolidate the text with what you currently have before lemma 2}
\vspace{-2mm}
\begin{proposition}\label{lm:first_order}
(First-order derivative of GCN~\cite{kang2022rawlsgcn}) Given an $L$-layer GCN whose parameters are $\Theta$, an input graph $\mathcal{G}=\{\mathcal{V}, \mathbf{A}, \mathbf{X}\}$, a node $i$ with its label $\mathbf{y}_i$ and a node-wise loss function $r(i, \mathbf{y}_i, \Theta)$ for node $i$, the first-order derivative of loss function $r(i, \mathbf{y}_i, \Theta)$ with respect to the parameters $\mathbf{W}^{(l)}$ in the $l$-th graph convolution layer is
\begin{equation}\label{eq:first_order}
    \nabla_{\mathbf{W}^{(l)}} r(i, \mathbf{y}_i, \Theta)
	= \big(\mathbf{\hat A}\mathbf{E}^{(l-1)}\big)^T
	  \bigg(\frac{\partial r(i, \mathbf{y}_i, \Theta)}{\partial \mathbf{E}^{(l)}} \circ  
		  \sigma'(\mathbf{\hat A} \mathbf{E}^{(l-1)} \mathbf{W}^{(l)})\bigg)
\end{equation}
where $\mathbf{\hat A} = \mathbf{\tilde D}^{-\frac{1}{2}} (\mathbf{A} + \mathbf{I}) \mathbf{\tilde D}^{-\frac{1}{2}}$ is the renormalized graph Laplacian with $\mathbf{\tilde D}$ being the degree matrix of $\mathbf{A} + \mathbf{I}$, $\sigma'$ is the derivative of the activation function $\sigma$, $\circ$ is the element-wise product, and $\frac{\partial r(i, \mathbf{y}_i, \Theta)}{\partial \mathbf{E}^{(l)}}$ can be iteratively calculated by
\begin{equation}\label{eq:first_order_wrt_embeddings}
    \frac{\partial r(i, \mathbf{y}_i, \Theta)}{\partial \mathbf{E}^{(l)}} = 
        \mathbf{\hat A}^T
        \bigg(
            \frac{\partial r(i, \mathbf{y}_i, \Theta)}{\partial \mathbf{E}^{(l+1)}}
            \circ
            \sigma'(\mathbf{\hat A} \mathbf{E}^{(l)} \mathbf{W}^{(l+1)})
        \bigg)
        \big(\mathbf{W}^{(l+1)}\big)^T
\end{equation}
\end{proposition}
\begin{proof}
See Appendix.
\end{proof}
\vspace{-3mm}
Since the parameters are often represented as matrices in the hidden layers, the second-order derivative will be a 4-dimensional tensor (i.e., a Hessian tensor). Building upon the results in Proposition~\ref{lm:first_order}, we first present the computation of the Hessian tensor in Theorem~\ref{lm:second_order}. Then we discuss efficient computation of influence function, which is summarized in Algorithm~\ref{alg:hvp}.
\vspace{-2mm}

\begin{theorem}\label{lm:second_order}
(The Hessian tensor of GCN) Following the settings of Proposition~\ref{lm:first_order}, denoting the overall loss $R(\mathcal{G}, \mathcal{Y}_{\textrm{train}}, \Theta)$ as $R$ and $\sigma'_l$ as $\sigma'(\mathbf{\hat A} \mathbf{E}^{(l-1)} \mathbf{W}^{(l)})$, the Hessian tensor $\mathfrak{H}_{l,i} = \frac{\partial^2 R}{\partial \mathbf{W}^{(l)} \partial \mathbf{W}^{(i)}}$ of $R$ with respect to $\mathbf{W}^{(l)}$ and $\mathbf{W}^{(i)}$ has the following forms.
\begin{itemize}[
    align=left,
    leftmargin=2em,
    itemindent=0pt,
    labelsep=0pt,
    labelwidth=1em,
]
    \item [\textbf{Case 1.}] $i = l$, $\mathfrak{H}_{l,i} = 0$
    % \begin{equation}\label{eq:second_order_i=l}
    %     \mathfrak{H}_{l,i} = 0
    % \end{equation}
    \item [\textbf{Case 2.}] $i = l-1$
    \begin{equation}\label{eq:second_order_i=l-1}
        \mathfrak{H}_{l, i}[:,:,c,d] = 
            \bigg(\mathbf{\hat A}\frac{\partial \mathbf{E}^{(l-1)}}{\partial \mathbf{W}^{(i)}[c,d]}\bigg)^T
            \bigg(
                \frac{\partial R}{\partial \mathbf{E}^{(l)}}
                \circ
                \sigma'_l
            \bigg)
    \end{equation}
    where $\frac{\partial \mathbf{E}^{(l-1)}}{\partial \mathbf{W}^{(i)}[c,d]}$ is the matrix whose entry at the $a$-th row and the $b$-th column is %form of Eq.~\eqref{eq:key_term_2} when $i=l-1$.
    \begin{equation}\label{eq:case_2_key_term}
        \frac{\partial \mathbf{E}^{(l-1)} [a, b]}{\partial \mathbf{W}^{(l-1)}[c, d]} = \sigma'_{l-1}[a, b] \big(\mathbf{\hat A} \mathbf{E}^{(l-2)}\big)[a, c] \mathbf{I}[b, d]
    \end{equation}
    \item [\textbf{Case 3.}] $i < l-1$
    \begin{itemize}[
        align=left,
        leftmargin=2em,
        itemindent=0pt,
        labelsep=0pt,
        labelwidth=1em,
    ]
        \item Apply Eq.~\eqref{eq:case_2_key_term} for the $i$-th hidden layer.
        \item Forward to the $(l-1)$-th layer iteratively with
        \begin{equation}\label{eq:second_order_forwarding}
            \frac{\partial \mathbf{E}^{(l-1)}}{\partial \mathbf{W}^{(i)}[c,d]} = \sigma'_{l-1} \circ \bigg(\mathbf{\hat A}\frac{\partial \mathbf{E}^{(l-2)}}{\partial \mathbf{W}^{(i)}[c,d]} \mathbf{W}^{(l-1)}\bigg)
        \end{equation}
        \item Apply Eq.~\eqref{eq:second_order_i=l-1}.
    \end{itemize}
    \item [\textbf{Case 4.}] $i = l+1$
    \begin{equation}\label{eq:second_order_i=l+1}
        \mathfrak{H}_{l,i}[:,:,c,d] = 
            (\mathbf{\hat A}\mathbf{E}^{(l-1)})^T 
            \bigg(
                \frac{\partial^2 R}{\partial \mathbf{E}^{(l)} \partial \mathbf{W}^{(i)}[c, d]}
                \circ
                \sigma'_l
            \bigg)
    \end{equation}
    where $\frac{\partial^2 R}{\partial \mathbf{E}^{(l)}[a,b] \partial \mathbf{W}^{(l+1)}[c, d]} = \mathbf{I}[b, c] \big[\mathbf{\hat A}^T \big(\frac{\partial R}{\partial \mathbf{E}^{(l+1)}} \circ \sigma'_{l+1}\big)\big][a, d]$. %$\mathbf{1}[a, d] = \mathbb{1}[a==d]$
    \item [\textbf{Case 5.}] $i > l+1$
    \begin{itemize}[
        align=left,
        leftmargin=2em,
        itemindent=0pt,
        labelsep=0pt,
        labelwidth=1em,
    ]
        \item Compute $\frac{\partial^2 R}{\partial \mathbf{E}^{(i-1)} \partial \mathbf{W}^{(i)}[c, d]}$ whose $(a,b)$-th entry has the form $\frac{\partial^2 R}{\partial \mathbf{E}^{(i-1)}[a,b] \partial \mathbf{W}^{(i)}[c, d]} = \mathbf{I}[b, c] \bigg(\mathbf{\hat A}^T \bigg(\frac{\partial R}{\partial \mathbf{E}^{(i)}} \circ \sigma'_i\bigg)\bigg)[a, d]$
        \item Backward to $(l+1)$-th layer iteratively with
        \begin{equation}\label{eq:second_order_backward}
            \frac{\partial^2 R}{\partial \mathbf{E}^{(l)} \partial \mathbf{W}^{(i)}[c, d]} = 
                \mathbf{\hat A}^T 
                \bigg(
                    \frac{\partial^2 R}{\partial \mathbf{E}^{(l+1)} \partial \mathbf{W}^{(i)}[c, d]}
                    \circ
                    \sigma'_{l+1}
                \bigg)
                \big(\mathbf{W}^{(l+1)}\big)^T
        \end{equation}
        \item Apply Eq.~\eqref{eq:second_order_i=l+1}.
    \end{itemize}
\end{itemize}
\end{theorem}
\begin{proof}
See Appendix.
\end{proof}

%A major challenge in computing the second-order derivative lies in its natural formulation as a high-dimensional tensor. 
  %We denote the Hessian tensor of loss function $R$
% \footnote{We use $R$ to represent $R(\mathcal{G}, \mathcal{Y}_{\textrm{train}}, \Theta)$ for notational simplicity.} 
% with respect to the parameters in $l$-th and $i$-th hidden layers as $\mathfrak{H}_{l,i}$, i.e., $\mathfrak{H}_{l,i} = \frac{\partial^2 R}{\partial \mathbf{W}^{(l)} \partial \mathbf{W}^{(i)}}$. It is trivial that the $(a,b,c,d)$-th entry of the Hessian tensor $\mathfrak{H}_{l,i}$ is equivalent to calculating the second-order derivative element-wise, i.e., $\mathfrak{H}_{l,i}[a,b,c,d] = \frac{\partial^2 R}{\partial \mathbf{W}^{(l)}[a,b] \partial \mathbf{W}^{(i)}[c,d]}$. With this observation, we first show how to compute the element-wise second-order derivative and then present our solution to avoid tensor computation based on the Hessian tensor (Algorithm~\ref{alg:hvp}). 

% \hh{from here to the end of sec 3.2., would it be better to move it to section 3.3 as part of the JuryGCN algorithm?}\jian{i think the following parts are still about influence function computation, it might be better to put together with the proposition and lemma. and i find it a bit weird to have the remark about potential homophily in the next section.}\hh{ok}
\vspace{-2mm}
Even with Proposition~\ref{lm:first_order} and Theorem~\ref{lm:second_order}, it is still non-trivial to compute the influence of node $u$ due to (C1) the high-dimensional nature of the Hessian tensor and (C2) the high computational cost of Eq.~\eqref{eq:node_influence} due to the inverse operation. Regarding the first challenge (C1), for any node $u$, we observe that each element in the first-order derivative $\nabla_{\mathbf{W}^{(l)}} R$ is the element-wise first-order derivative, i.e., $\nabla_{\mathbf{W}^{(l)}} R[a, b] = \frac{\partial R}{\partial \mathbf{W}^{(l)}[a,b]}$. Likewise, for the Hessian tensor, we have $\mathfrak{H}_{l,i}[a,b,c,d] = \frac{\partial^2 R}{\partial\mathbf{W}^{(l)}[a,b] \partial\mathbf{W}^{i}[c,d]}$.\footnote{We use $R$ to represent $R(\mathcal{G}, \mathcal{Y}_{\textrm{train}}, \Theta)$ for notational simplicity.} Thus, the key idea to solve the first challenge (C1) is to vectorize the first-order derivative into a column vector and compute the element-wise second-order derivatives accordingly, which naturally flatten the Hessian tensor into a  %flattened 
Hessian matrix. More specifically, we first vectorize the first-order derivatives of $R$ with respect to $\mathbf{W}^{(l)},\forall l\in\{1,\ldots, L\}$ in to column vectors and stack them vertically as follows,
\begin{equation}\label{eq:vectorize_first_order}
    \mathbf{f}_{R} = 
    \left[\begin{array}{c}
        \textrm{vec}(\nabla_{\mathbf{W}^{(1)}} R) = \textrm{vec}(\frac{\partial R}{\partial \mathbf{W}^{(1)}}) \\
        \vdots \\
        \textrm{vec}(\nabla_{\mathbf{W}^{(L)}} R) = \textrm{vec}(\frac{\partial R}{\partial \mathbf{W}^{(L)}})
    \end{array}\right]
\end{equation}
where $\textrm{vec}()$ vectorizes a matrix to a column vector. Then the flattened Hessian matrix is a matrix $\mathbf{H}_{\textrm{flat}}$ whose rows are of the form
\begin{equation}\label{eq:flattened_hessian_matrix}
    \mathbf{H}_{\textrm{flat}}[(i\cdot c + d), :] = \bigg(\frac{\partial \mathbf{f}_R}{\partial \mathbf{W}^{(i)}[c,d]}\bigg)^T = \textrm{vec}(\mathfrak{H}_{l,i}[:,:,c,d])^T
\end{equation}
Finally, we follow the strategy to compute the influence of node $u$: (1) Compute $\nabla_{\mathbf{W}^{(l)}} r(u, \mathbf{y}_u, \Theta)$ for all $l$-th hidden layer; (2) Vectorize $\nabla_{\mathbf{W}^{(l)}} r(u, \mathbf{y}_u, \Theta)$ and stack to column vector $\mathbf{f}_u$ as shown in Eq.~\eqref{eq:vectorize_first_order}; (3) Compute the influence function $\mathbb{I}(u) = \mathbf{H}_{\textrm{flat}}^{-1} \mathbf{f}_u$.

Regarding the second challenge (C2), the key idea is to apply Hessian-vector product (Algorithm~\ref{alg:hvp})~\cite{koh2017understanding, alaa2020discriminative}, which approximates $\mathbb{I}(u) = \mathbf{H}_{\textrm{flat}}^{-1} \mathbf{f}_u$ using the power method. Mathematically, it treats $\mathbf{H}_{\textrm{flat}}^{-1} \mathbf{f}_u$ as one vector and iteratively computes 
\vspace{-1mm}
% $\mathbf{H}_{\textrm{flat}}^{-1}\mathbf{f}_u = \mathbf{f}_u + (\mathbf{I} - \mathbf{\hat H}_{\textrm{flat}}) \big(\mathbf{H}_{\textrm{flat}}^{-1}\mathbf{f}_u\big)$
\begin{equation}\label{eq:hvp_recursion}
    \mathbf{H}_{\textrm{flat}}^{-1}\mathbf{f}_u = \mathbf{f}_u + (\mathbf{I} - \mathbf{\hat H}_{\textrm{flat}}) \big(\mathbf{H}_{\textrm{flat}}^{-1}\mathbf{f}_u\big)
\end{equation}
where $\mathbf{\hat H}_{\textrm{flat}}\big(\mathbf{H}_{\textrm{flat}}^{-1}\mathbf{f}_u\big)$ is viewed as a vector and $\mathbf{\hat H}_{\textrm{flat}}$ is the flattened Hessian matrix with respect to a set of sampled nodes at current iteration. The workflow of the Hessian-vector product is presented in Algorithm~\ref{alg:hvp}. For any $l$-th hidden layer (step 2), we first compute the first-order derivative $\nabla_{\mathbf{W}^{(l)}}r(u, \mathbf{y}_u, \Theta)$ with respect to $\mathbf{W}^{(l)}$ (step 3). Then we vectorize it to a column vector and stack it to $\mathbf{f}_u$ that stores the first-order derivatives of all hidden layers (step 4). After all first-order derivatives are computed, we apply the power method to compute the Hessian-vector product. In each iteration, we first sample a batch of $t$ training nodes, which helps reduce both noise and running time, and then compute the empirical loss over these nodes (steps 7 -- 8). After that, we compute the second-order derivatives with Theorem~\ref{lm:second_order} and flatten it to a matrix with the strategy shown in Eq.~\eqref{eq:flattened_hessian_matrix} (step 9). We finish this iteration by computing Eq.~\eqref{eq:hvp_recursion} (step 10). The power method (steps 7 -- 10) iterates until the maximum number of iteration is reached to ensure the convergence. Consequently, Algorithm~\ref{alg:hvp} offers a computationally friendly way to approximate influence functions without involving both tensor-level operations and the computationally expensive matrix inversion.
% \vspace{-3mm}
\begin{algorithm}[t]
    \SetKwInOut{Input}{Input}
    \SetKwInOut{Output}{Output}
    \Input{An input graph $\mathcal{G}$, training nodes $\mathcal{V}_{\textrm{train}}$, ground-truth labels $\mathcal{Y}_{\textrm{train}}$, node $u$ with label $\mathbf{y}_u$, an $L$-layer GCN with parameters $\Theta$, a node-wise loss function $r$, sampling batch size $t$, \#iterations $m$;}
	\Output{The influence $\mathbb{I}_{\Theta}(u)$ of node $u$.}
	
	Initialize $\mathbf{f}_u = []$ as an empty column vector\;
	\For{$l = 1 \rightarrow L$}{
	    Compute $\nabla_{\mathbf{W}^{(l)}} r(u, \mathbf{y}_u, \Theta)$ by Eq.~\eqref{eq:first_order}\;
	    Vectorize $\nabla_{\mathbf{W}^{(l)}} r(u, \mathbf{y}_u, \Theta)$ and stack it to $\mathbf{f}_u$ as Eq.~\eqref{eq:vectorize_first_order}\;
	}
	
	Initialize $\big(\mathbf{H}_{\textrm{flat}}^{-1}\mathbf{f}_u\big)_0 \leftarrow \mathbf{f}_u$\;
	\For{$\textit{iter} = 1 \rightarrow \textit{m}$}{
	    Uniformly sample $t$ training nodes and get $\mathcal{V}_{\textrm{s}}$\;
	    Compute empirical loss $R_s \leftarrow \frac{1}{|\mathcal{V}_{\textrm{s}}|}\sum_{i\in\mathcal{V}_{\textrm{s}}} r(i, \mathbf{y}_i, \Theta)$\;
	    Compute $\mathbf{\hat H}_{\textrm{flat}}$ of $R_s$ with Theorem~\ref{lm:second_order} and Eq.~\eqref{eq:flattened_hessian_matrix}\;
	    Compute $\big(\mathbf{H}_{\textrm{flat}}^{-1}\mathbf{f}_u\big)_{\textit{iter}} \leftarrow \mathbf{f}_u + (\mathbf{I} - \mathbf{\hat H}_{\textrm{flat}}) \big(\mathbf{H}_{\textrm{flat}}^{-1}\mathbf{f}_u\big)_{\textit{iter}-1}$
	}
	\Return $\big(\mathbf{H}_{\textrm{flat}}^{-1}\mathbf{f}_u\big)_{m}$\;
	
    \caption{Hessian-Vector Product}
    \label{alg:hvp}
    
\end{algorithm}

\noindent \textit{Remark.} We observe that both the first-order derivative (Proposition~\ref{lm:first_order}) and the second-order derivative (Theorem~\ref{lm:second_order}) can be computed in the style of neighborhood aggregation. Due to the well-known over-smoothness of GCN and the homophily nature of neighborhood aggregation, the resulting influence functions by Algorithm~\ref{alg:hvp} may follow the homophily principle as well. For two nodes under homophily, due to similarity between their influences, their corresponding LOO parameters and LOO errors could be similar as well, which in turn could cause similar uncertainty scores by Definition~\ref{defn:jackknife_uncertainty}. The potential homophily phenomenon in jackknife uncertainty is consistent with the homophily assumption with respect to uncertainty/confidence in existing works~\cite{zhao2020uncertainty, stadler2021graph, wang2021confident}.
% \hh{for either alg 1 and/or alg 2, do we have some complexity analysis?}\jian{no complexity analysis for now. it deals with multiple backward and forward propagation through hidden layers since the hession tensor has many cases to consider. }\hh{with the runtime comparison in sec 5, i think we should be fine}

\vspace{-3mm}
\section{\method: Algorithm and Applications}\label{sec:algorithm}
In this section, we present our proposed method named \method\ to quantify node jackknife uncertainty (Algorithm~\ref{alg:uq}) followed by discussions on applications and generalizations of \method.
\vspace{-4mm}
\subsection{\method\ Algorithm }\label{subsec:algorithm}
\vspace{-1mm}
% \hh{if we put this as a separate section, subsectin 4.1: jurygcn algorithms (algoirthm 2 (and maybe algorithm 1)), subsection 4.2: jurygCN applications.}
With Algorithm~\ref{alg:hvp}, the LOO parameters of each node can be efficiently computed by proper initialization on the perturbation coefficient ($\epsilon$ in Eq.~\eqref{eq:node_influence}). After that, the LOO predictions and LOO errors can be efficiently inferred by simply switching the original parameters to the LOO parameters, resulting in efficient jackknife uncertainty quantification. 

Based on that, Algorithm~\ref{alg:uq} presents the general workflow of our proposed \method\ to quantify the jackknife uncertainty of a node. In detail, with proper initialization (step 1), we loop through each training node $i$ to quantify their influences (step 2). For each training node $i$, it estimates the LOO parameters by leaving out training node $i$ (steps 3 -- 6), outputs the LOO predictions of nodes $i$ and $u$ (step 7) and compute the LOO error of each training node $i$ (step 8). After the LOO predictions of node $u$ and the LOO errors of all training nodes are obtained, we compute the lower bound and upper bound of the predictive confidence interval (steps 9 -- 10). Finally, the uncertainty of the node is computed as the width of the predictive confidence interval (step 11).
\vspace{-4mm}
\begin{algorithm}
	\SetKwInOut{Input}{Input}
	\SetKwInOut{Output}{Output}
	\Input{An input graph $\mathcal{G} =\{\mathcal{V}, \mathbf{A}, \mathbf{X}\}$ with training nodes $\mathcal{V}_{\textrm{train}}$, a node $u$, a GCN with parameters $\Theta$, a node-wise loss function $r$, a coverage parameter $\alpha$;}
	\Output{The uncertainty $\mathbb{U}_\Theta(u)$ of node $u$.}
	Initialize $\epsilon \leftarrow -\frac{1}{|\mathcal{V}_{\textrm{train}}|}$\;
% 	Output vanilla prediction $\textit{GCN}(u, \Theta)$\;\jian{double check if we can drop it}
	\For{$i \in \mathcal{V}_{\textrm{train}}$}{
    % 	Output vanilla prediction $\textit{GCN}(i, \Theta)$\;\jian{double check if we can drop it}
    	Compute node-wise loss $r_{i, \Theta} \leftarrow r\big(i, \mathbf{y}_i, \Theta\big)$\;%\qh{according to Eq.(5), is the second term ($\textit{GCN}(i, \Theta)$) label, i.e., $r\big(i, y_i, \Theta\big)$?}\jian{good catch! fixed.}
    	Compute node-wise derivative $\nabla_{\Theta} r_{i, \Theta}$\;
    	Compute $\mathbb{I}_{\Theta}(i) \leftarrow \mathbf{H}_{\Theta}^{-1}\nabla_{\Theta} r_{i, \Theta}$ using Algorithm~\ref{alg:hvp}\;
    	Estimate LOO model parameters $\Theta_{\epsilon, i} \leftarrow \Theta + \epsilon \mathbb{I}_{\Theta}(i)$\;
    	Output LOO predictions $\textit{GCN}(i, \Theta_{\epsilon, i})$ and $\textit{GCN}(u, \Theta_{\epsilon, i})$\;
    	Compute LOO error $\textit{err}_i \leftarrow \|\mathbf{y}_i - \textit{GCN}(i, \Theta_{\epsilon, i})\|_2$\;
	}
	Compute lower bound $\mathbb{C}_{\Theta}^{-}(u) \leftarrow Q_{\alpha}(\{\|\textit{GCN}(u, \Theta_{\epsilon, i})\|_2 - \textit{err}_i | i\in\mathcal{V}_{\textrm{train}}\})$\;
	Compute upper bound $\mathbb{C}_{\Theta}^{+}(u) \leftarrow Q_{1-\alpha}(\{\|\textit{GCN}(u, \Theta_{\epsilon, i})\|_2 + \textit{err}_i | i\in\mathcal{V}_{\textrm{train}}\})$\;
	\Return $\mathbb{U}_\Theta(u) \leftarrow \mathbb{C}_{\Theta}^{+}(u) - \mathbb{C}_{\Theta}^{-}(u)$\;
	\caption{\method: {\small Jackknife Uncertainty Quantification}}
	\label{alg:uq}
\end{algorithm}
% \vspace{-2mm}

\subsection{\method\ Applications}
\label{subsec:application}
\vspace{-1mm}
After quantifying the jackknife uncertainty of each node, we utilize the node uncertainty in (1) active learning on node classification and (2) semi-supervised node classification. The details of uncertainty-aware active learning and node classification are as follows.

\noindent \textbf{Application \# 1: Active Learning on Node Classification.} In general, active learning sequentially selects a subset of data points to query according to an acquisition function, which is designed to identify the most informative samples, and hence improves the model performance from the obtained labels. In active learning on node classification, we are given (1) an unlabelled training set of nodes (i.e., $\mathcal{V}_{\text{train}}$), (2) a node classifier (e.g., GCN), (3) step size $b$, and (4) the query budget $K$. At each query step, according to the acquisition function, we select $b$ nodes from the remaining unlabelled nodes in the training set $\mathcal{V}_{\mathrm{train}}$ to query and then re-train the classifier. The query step is repeated until the query budget $K$ is exhausted. 
Intuitively, a node with high predictive uncertainty is a better query candidate compared to the one with certain prediction. From Algorithm~\ref{alg:uq}, we can obtain the jackknife uncertainty of each node in $\mathcal{V}_{\mathrm{train}}$, hence, we define the acquisition function as follows, $\mathrm{Acq}(\mathcal{V}_{\mathrm{train}})=\argmax_{u\in\mathcal{V}_{\mathrm{train}}} \mathbb{U}_\Theta(u),$
% \begin{equation}
%     \mathrm{Acq}(\mathcal{V}_{\mathrm{train}})=\argmax_{u\in\mathcal{V}_{\mathrm{train}}} \mathbb{U}_\Theta(u)
% \end{equation} \jian{make it an inline equation?}
where $\mathbb{U}_\Theta(u)$ is the jackknife uncertainty of node $u$ from the remaining unlabelled nodes in $\mathcal{V}_{\mathrm{train}}$ and is computed in Eq.~\eqref{eq:jackknife_uncertainty}. Therefore, at each step, we select $b$ unlabelled nodes with the top-$b$ jackknife uncertainty. The detailed experimental settings are introduced in Appendix.
% pool for query -> acquisition function (detailed setup in section 4)

\noindent \textbf{Application \# 2: Semi-supervised Node Classification.} In training a GCN-based node classification model, existing approaches treat each training node equally and compute the mean of loss from all training nodes, i.e., $R=\frac{1}{|\mathcal{V}_{\textrm{train}}|} \sum_{i\in\mathcal{V}_{\textrm{train}}}r(i, \mathbf{y}_i, \Theta)$ where the node-specific loss (i.e., $r(\cdot)$) is factored by an identical weight (i.e., $\frac{1}{|\mathcal{V}_{\textrm{train}}|}$). Intuitively, training nodes have various levels of uncertainty during training, the easily predicted samples (i.e., small uncertainty) may comprise the majority of the total loss and hence dominate the gradient, which makes the training inefficient. To address this problem, based on jackknife uncertainty estimation, we introduce a dynamic scale factor, $\beta$, to adjust the importance for nodes with different levels of uncertainty.
% (1) up-weight nodes with larger uncertainty (i.e., difficult training samples), and (2) down-weight nodes with smaller uncertainty (i.e., easy samples).
Specifically, given the cross-entropy loss that is utilized in semi-supervised node classification, we define the uncertainty-aware node-specific loss as, %follows,
$r_u = - \beta_u^\tau\log(p^{(i)}_u)$
% \begin{equation}
%     r_u = - \beta_u^\tau\log(p^{(i)}_u)\label{eq:uncertainty-based_loss}
% \end{equation}
\noindent where $p^{(i)}_u$ is the predictive probability of the $i$-th class from GCN and $\tau$ is a hyperparameter. The scale factor of node $u$ is computed by normalizing the uncertainty over all training nodes, i.e., $\beta_u=\frac{|\mathbb{U}_\Theta(u)|}{\sqrt{\sum_{i\in\mathcal{V}_{\mathrm{train}}}|\mathbb{U}_\Theta(i)|^2}}$. By introducing $\beta_u$, %we can observe that 
the loss of node with larger uncertainty (i.e., $\mathbb{U}_\Theta$) would be upweighted in the total loss and hence the training is guided toward nodes with high uncertainty.
Therefore, when training a node classification model, we leverage Algorithm~\ref{alg:uq} to estimate the jackknife uncertainty of the training nodes and then apply the obtained uncertainty results to update the loss every few number of epochs.

In addition to Tasks 1 and 2, our proposed \method\ is generalizable to other learning tasks and graph mining models. Due to the space limit, we only present brief descriptions of each generalization direction, which could be a future direction of \method. %applying the proposed framework.

\noindent \textbf{Applications beyond Node Classification.} The proposed \method\ can be applied to a variety of learning tasks. For example, it can estimate the confidence interval of the predictive dependent variable in a regression task by replacing 
% and one direct modification is to replace 
the cross-entropy loss for node classification with mean squared error (MSE)~\cite{alaa2020discriminative}. Besides active learning, reinforcement learning (RL) is extensively explored to model the interaction between agent and environment. The proposed framework is applicable to RL in the following two aspects. First, in the early stage of training an RL model, the uncertainty quantification results can be utilized to guide exploration. Second, through effectively quantifying the uncertainty of the reward after taking certain actions, \method~ can also benefit the exploitation. Specifically, as a classic topic of RL, multi-armed bandit can also be a potential application for \method~ where uncertainty is highly correlated with decision making. %\hh{interesting. might worth further study for another paper}

% regression: loss from cross-entropy to MSE

% reinforcement learning: uncertainty as a criterion for making actions or reward (reducing uncertainty)

% multi-armed bandit: confidence interval

\noindent \textbf{Beyond \method: Generalizations to other GNN Models.} The proposed \method\ is able to be generalized to other graph mining models with the help of automatic differentiation in many existing packages, e.g., PyTorch, TensorFlow. In this way, only model parameters and the loss function are required in computing the first-order and second-order derivatives for influence function computation (Algorithm~\ref{alg:hvp}), which is further used in Algorithm~\ref{alg:uq} for jackknife uncertainty quantification.
\vspace{-3mm}
\section{Experimental Evaluation}~\label{sec:experiment}
In this section, we conduct experiments to answer the following research questions:
% In this section, we conduct experiments to demonstrate the effectiveness of the proposed framework. To be specific, we aim to answer the following research questions:
\begin{itemize}[
    align=left,
    leftmargin=2em,
    itemindent=0pt,
    labelsep=2pt,
    labelwidth=1em,
]
    \item [\textbf{RQ1.}] How effective is \method~ in improving GCN predictions?
    \item [\textbf{RQ2.}] How efficient is \method~ in terms of time and space?
    \item [\textbf{RQ3.}] How sensitive is \method~ to hyperparameters? 
\end{itemize}
\vspace{-3mm}
\subsection{Experimental Settings}
\label{subsec:exp_settings}
\noindent\textbf{1 -- Datasets.} We adopt four widely used benchmark datasets, including Cora~\cite{sen2008collective}, Citeseer~\cite{sen2008collective}, Pubmed~\cite{namata2012query} and Reddit~\cite{hamilton2017inductive}. %\jian{(1) maybe remove descriptions? they should be pretty well known (2) do we need 2 citations for cora and citeseer?} In the first three citation datasets, each network is a directed network where each node represents a document and an edge is a citation link, representing document A cites document B, or vice-versa with a direction. The node labels represent the different fields that the documents belong to. For Reddit dataset, the posts are regarded as nodes and the post-to-post network is constructed by connecting the posts that are commented by the same user. The node labels correspond to the different communities of the posts. 
Table~\ref{tab:data_stats} summarizes the statistics of the datasets.
\vspace{-3mm}
\begin{table}[!ht]
	\centering
% 	\vspace{-3mm}
	\caption{Statistics of datasets.}
% 	\vspace{-4mm}
    \scalebox{0.85}
	{\begin{tabular}{lcccc}
    \toprule
    \textbf{Datasets} & Cora  & Citeseer & PubMed & Reddit \\
    \midrule
    \# nodes & $2,708$ & $3,327$ & $19,717$ & $232,965$ \\
    \# edges & $5,429$ & $4,732$ & $44,338$ & $114,615,892$  \\
    \# features & $1,433$ & $3,703$ & $500$ & $602$  \\
    \# classes & $7$ & $6$ & $3$ & $41$ \\
    \bottomrule
	\end{tabular}
	}
% 	\vspace{-3mm}
	\label{tab:data_stats}
\end{table}
% \qh{description can be simplified or move to appendix.}
\vspace{-2mm}

\noindent\textbf{2 -- Comparison Methods.} We compare the proposed algorithm with node classification methods in the following two categories, including (1) active learning-based approaches, which aim to select the most informative nodes to query, and (2) semi-supervised methods. In active learning on node classification, we include the following methods for comparison, AGE~\cite{cai2017active}, ANRMAB~\cite{gao2018active}, Coreset~\cite{sener2017active}, Centrality, Random and SOPT-GCN~\cite{ng2018bayesian}. For semi-supervised node classification, we have  two uncertainty-based approaches, including S-GNN~\cite{zhao2020uncertainty} and GPN~\cite{stadler2021graph}, as well as GCN~\cite{kipf2017semi} and GAT~\cite{velivckovic2018graph}. For detailed description of the baselines, please refer to Appendix.

\noindent \textbf{3 -- Evaluation Metric.} In this paper, we use Micro-F1 to evaluate the effectiveness. In terms of efficiency, we compare the running time (in seconds) and the memory usage (in MB).

\noindent \textbf{4 -- Implementation Details.} We introduce the implementation details of the experiments in Appendix. 

\noindent \textbf{5 -- Reproducibility.} All datasets are publicly available. All codes are programmed in Python 3.6.9 and PyTorch 1.4.0. All experiments are performed on a Linux server with 96 Intel Xeon Gold 6240R CPUs and 4 Nvidia Tesla V100 SXM2 GPUs with 32 GB memory.
\vspace{-4mm}
\subsection{Effectiveness Results (RQ1)}\label{subsec:effectiveness_results}
\vspace{-1mm}
\noindent\textbf{1 -- Active Learning on Node Classification.} Table~\ref{tab:results_al} presents the results of active learning at different query steps. We highlight the best performing approach in bold and underline the best competing one respectively. We have the following observations: \textbf{(1)} At the final query step (i.e., the $5$-th line for each dataset), \method\ selects more valuable training nodes than other baseline methods, resulting in higher Micro-F1. Though AGE outperforms \method\ on \textit{Pubmed} at $40$ queries, the superiority is very marginal (i.e., $0.1\%$); \textbf{(2)} In general, \method\ achieves a much better performance when the query size is small. For example, on \textit{Cora} dataset, \method\ outperforms SOPT-GCN by $2.3\%$ at 20 queries while the gain becomes $0.6\%$ at 80 queries. One possible explanation is that newly-selected query nodes may contain less fruitful information about new classes/features for a GCN classifier that is already trained with a certain number of labels; \textbf{(3)} As the query size increases, the gained performance of each method is diminishing, which is consistent with the second observation. For instance, on \textit{Reddit} dataset, ANRMAB improves $13.5\%$ when query size becomes 100 (from 50), while the gain is only $1.5\%$ at 250 queries.

\begin{table*}[!ht]
\centering
% \small
\caption{Performance comparison results on active learning on node classification w.r.t. Micro-F1\hide{ on four datasets}. Higher is better.}
% \vspace{-4mm}
\scalebox{0.85}{
\begin{tabular}{cccccccccc}
\toprule
\textbf{Data} & \textbf{Query size} & \textbf{\method\ (Ours)} & \textbf{ANRMAB} & \textbf{AGE}  & \textbf{Coreset} & \textbf{Centrality} & \textbf{Degree} & \textbf{Random} & \textbf{SOPT-GCN} \\
\midrule
\multirow{5}{*}{Cora}   & 20 & $\bf 51.1\pm1.2$ & $46.8\pm0.5$ & \underline{$49.4\pm1.0$} & $43.8\pm0.8$ & $41.9\pm0.6$ & $38.5\pm0.7$ & $40.5\pm1.6$ & $48.8\pm0.7$ \\
                        & 40 & $\bf 64.7\pm0.8$ & $61.2\pm0.8$ & $58.2\pm0.7$ & $55.4\pm0.5$ & $57.3\pm0.7$ & $48.4\pm0.3$ & $56.8\pm1.3$ & \underline{$62.6\pm0.8$}\\
                        & 60 & $\bf 69.9\pm0.9$ & $67.8\pm0.7$ & $65.7\pm0.8$ & $62.2\pm0.6$ & $63.1\pm0.5$ & $58.8\pm0.6$ & $64.5\pm1.5$ & \underline{$67.9\pm0.6$}\\
                        & 80 & $\bf 74.2\pm0.7$ & $73.3\pm0.6$ & $72.5\pm0.4$ & $70.2\pm0.5$ & $69.1\pm0.4$ & $67.6\pm0.4$ & $69.7\pm1.6$ & \underline{$73.6\pm0.5$}\\
                        & 100 & $\bf 75.5\pm0.6$ & $74.9\pm0.4$ & $74.2\pm0.3$ & $73.8\pm0.4$ & $74.1\pm0.3$ & $73.0\pm0.2$ & $74.2\pm1.2$ & $\bf 75.5\pm0.7$\\
\midrule
\multirow{5}{*}{Citeseer} & 20 & $\bf 38.4\pm1.5$ & $35.9\pm1.0$ & $33.1\pm0.9$ & $30.2\pm1.2$ & $35.6\pm1.1$ & $31.5\pm0.9$ & $30.3\pm2.3$ & \underline{$36.1\pm0.7$}\\
                          & 40 & $\bf 51.1\pm0.9$ & $46.7\pm1.3$ & $49.5\pm0.6$ & $42.1\pm0.8$ & \underline{$49.8\pm1.3$} & $39.8\pm0.7$ & $41.1\pm1.8$ & $49.2\pm0.5$\\
                          & 60 & $\bf 58.2\pm0.8$ & $55.2\pm0.9$ & $56.1\pm0.5$ & $52.1\pm0.9$ & \underline{$57.1\pm0.7$} & $50.1\pm1.1$ & $49.8\pm1.3$ & $56.4\pm0.5$\\
                          & 80 & $\bf 63.8\pm1.1$ & \underline{$63.2\pm0.7$} & $61.5\pm0.8$ & $59.9\pm0.6$ & $63.3\pm1.0$ & $58.8\pm0.6$ & $58.1\pm1.1$ & $63.2\pm0.8$\\
                          & 100 & $\bf 64.3\pm1.2$ & \underline{$64.1\pm0.5$} & $63.2\pm0.7$ & $62.8\pm0.4$ & $63.9\pm0.6$ & $61.8\pm0.5$ & $62.9\pm0.8$ & $63.8\pm0.6$\\
\midrule
\multirow{5}{*}{Pubmed}  & 10 & $\bf 61.8\pm0.9$ & $60.5\pm1.3$ & $58.9\pm1.1$ & $53.1\pm0.7$ & $55.8\pm1.2$ & $56.4\pm1.5$ & $52.4\pm1.7$ & $59.5\pm0.6$\\
                         & 20 & $\bf 70.2\pm0.6$ & $66.8\pm1.1$ & \underline{$68.7\pm0.7$} & $62.8\pm0.5$ & $67.2\pm1.4$ & $64.3\pm1.0$ & $60.5\pm1.4$ & $67.9\pm0.9$\\
                         & 30 & $\bf 73.9\pm0.3$ & $71.6\pm0.8$ & \underline{$72.8\pm1.0$} & $68.9\pm0.3$ & $73.5\pm0.9$ & $70.1\pm0.7$ & $68.9\pm1.1$ & $72.3\pm0.8$\\
                         & 40 & \underline{$74.6\pm0.4$} & $73.2\pm0.6$ & $\bf 74.7\pm0.8$ & $72.8\pm0.8$ & $74.1\pm0.7$ & $72.0\pm0.8$ & $71.8\pm1.2$ & $73.8\pm0.7$\\
                         & 50 & $\bf 75.4\pm0.5$ & $74.7\pm0.4$ & $75.1\pm0.5$ & $73.5\pm0.6$ & $74.2\pm0.6$ & $72.9\pm0.5$ & $73.1\pm1.0$ & \underline{$75.2\pm0.5$}\\
\midrule
\multirow{5}{*}{Reddit}  & 50 & $\bf 69.7\pm1.7$ & $67.8\pm0.9$ & $64.2\pm1.1$ & $62.1\pm0.6$ & $65.5\pm1.2$ & $62.5\pm1.4$ & $63.7\pm2.4$ & \underline{$68.1\pm1.2$} \\
                         & 100 & $\bf 82.9\pm1.5$ & \underline{$81.3\pm1.0$} & $79.5\pm0.8$ & $81.2\pm1.0$ & $78.2\pm0.9$ & $81.1\pm1.2$ & $80.5\pm1.6$ & $80.4\pm1.3$\\
                         & 150 & $\bf 86.0\pm1.4$ & $84.3\pm0.7$ & $83.2\pm0.4$ & $84.8\pm0.9$ & $84.1\pm1.1$ & $82.5\pm1.2$ & $81.5\pm1.4$ & \underline{$85.0\pm1.5$}\\
                         & 200 & $\bf 88.1\pm0.9$ & $86.1\pm0.8$ & $85.8\pm0.5$ & $85.5\pm0.8$ & $87.5\pm0.8$ & $85.4\pm0.7$ & $83.1\pm1.8$ & \underline{$87.2\pm0.9$}\\
                         & 250 & $\bf 89.2\pm0.8$ & $87.6\pm0.7$ & $87.1\pm0.4$ & $86.6\pm1.1$ & \underline{$88.7\pm0.6$} & $86.1\pm1.0$ & $87.3\pm1.5$ & $87.8\pm1.1$\\
\bottomrule
\end{tabular}
}
\vspace{-2mm}
\label{tab:results_al}
\end{table*}

\begin{figure*}[!ht]
    \centering
    \scalebox{0.95}{
    \begin{subfigure}{.24\linewidth}
    \centering
    \includegraphics[width=.98\linewidth]{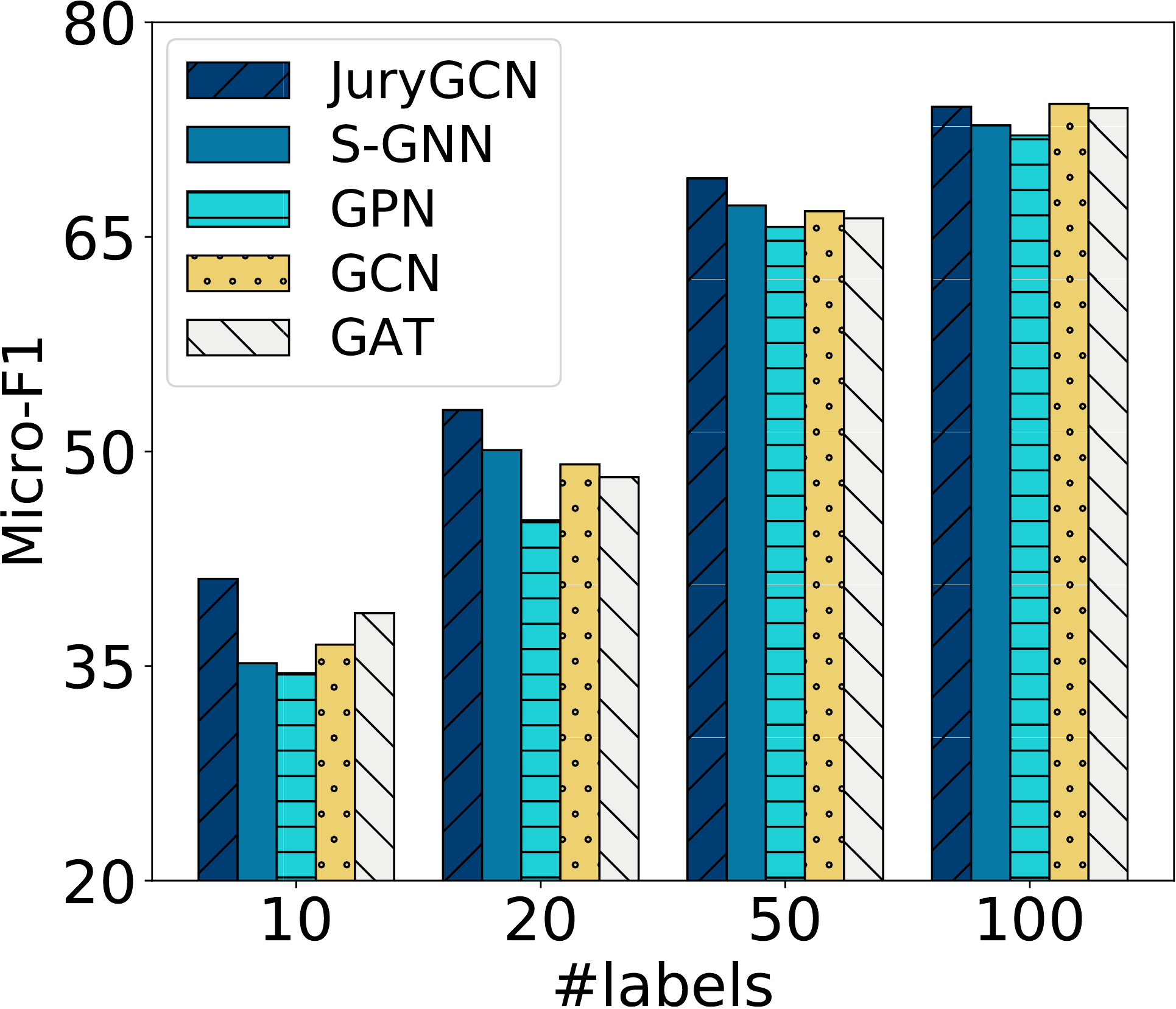}
    \vspace{-2mm}
    \caption{Cora}
    \label{fig:results_fsl_cora}
    \end{subfigure}%
    \begin{subfigure}{.24\linewidth}
    \centering
    \includegraphics[width=.98\linewidth]{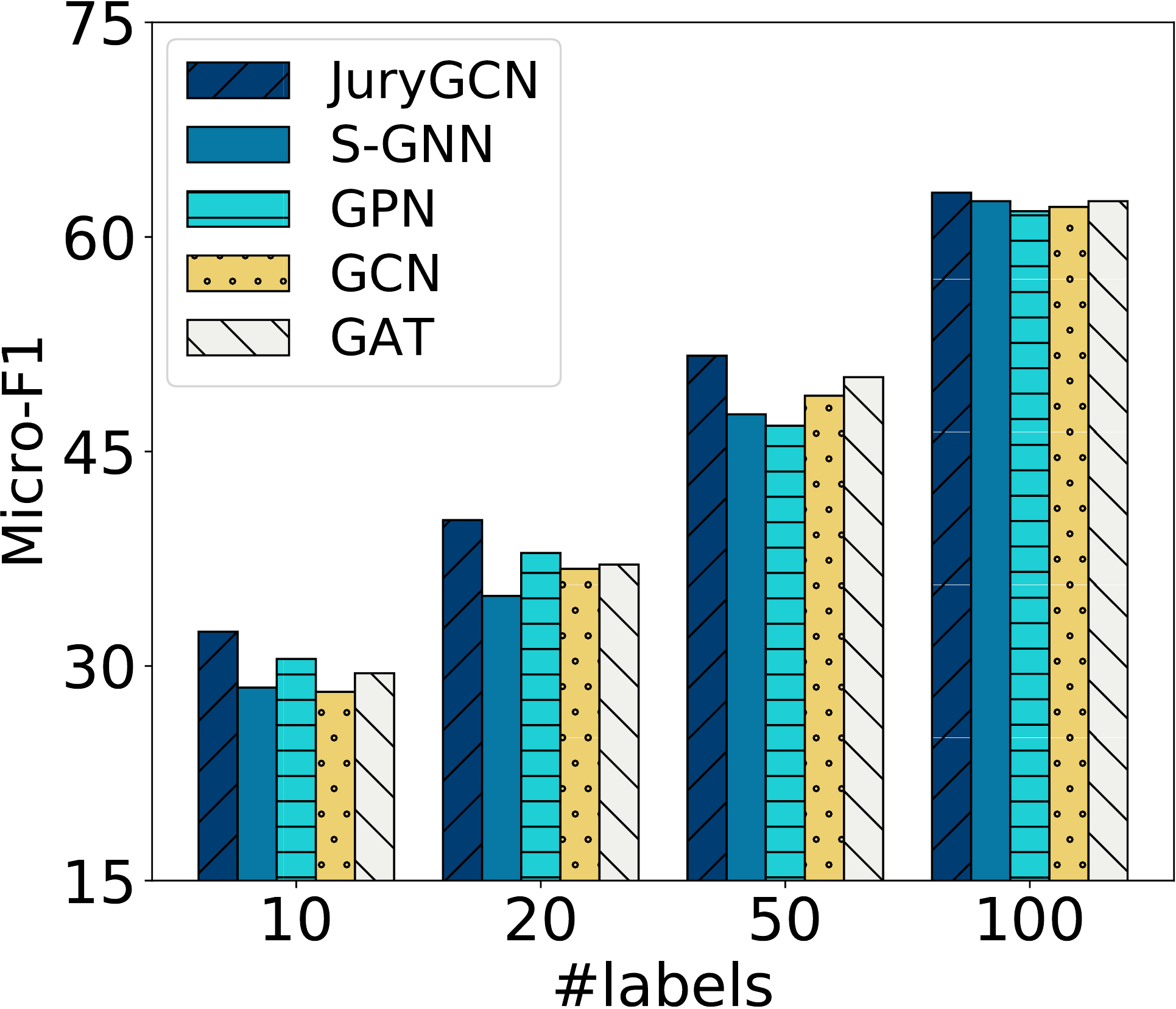}
    \vspace{-2mm}
    \caption{Citeseer}
    \label{fig:results_fsl_citeseer}
    \end{subfigure}
    \begin{subfigure}{.24\linewidth}
    \centering
    \includegraphics[width=.98\linewidth]{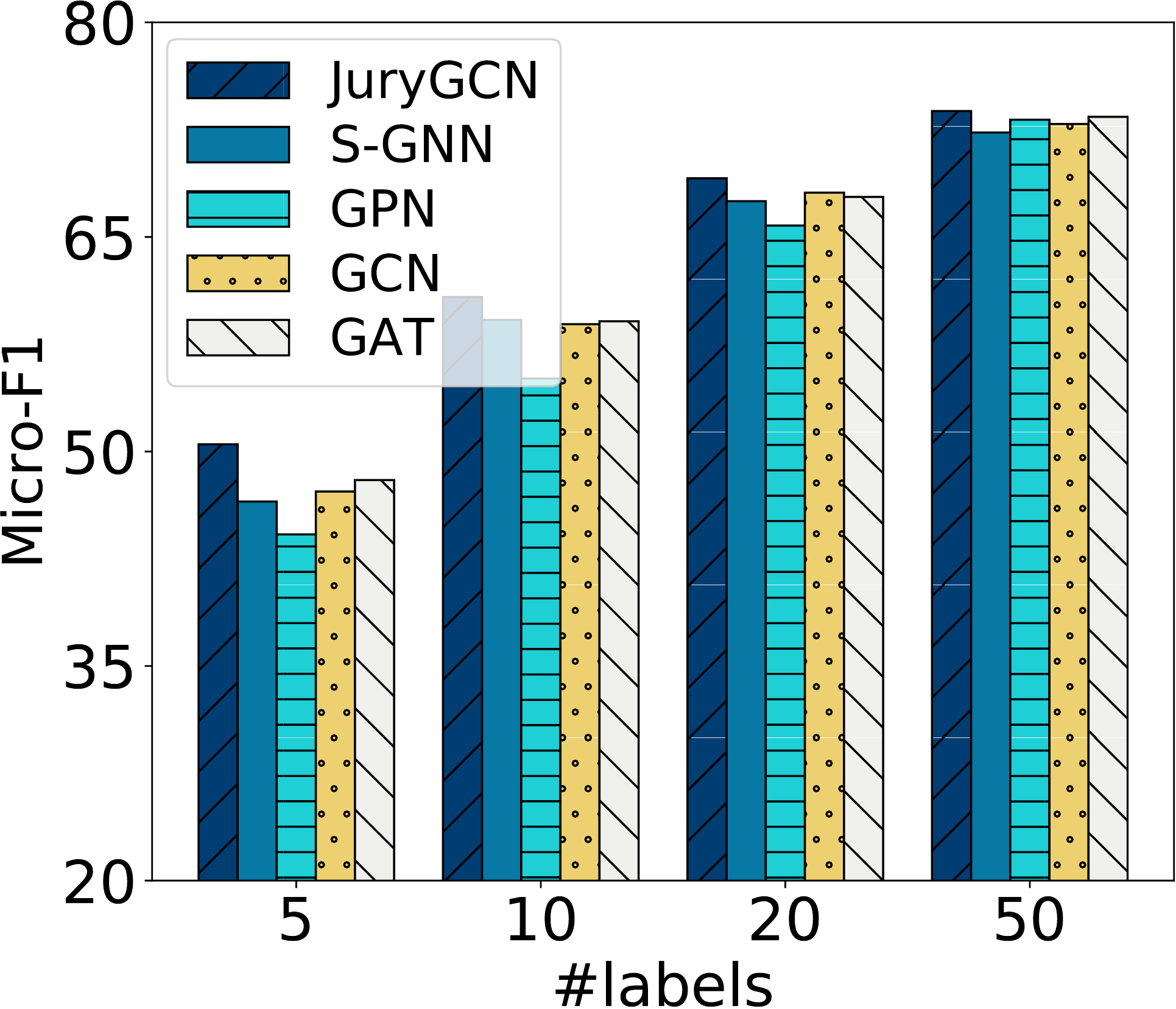}
    \vspace{-2mm}
    \caption{Pubmed}
    \label{fig:results_fsl_pubmed}
    \end{subfigure}
    \begin{subfigure}{.24\linewidth}
    \centering
    \includegraphics[width=.98\linewidth]{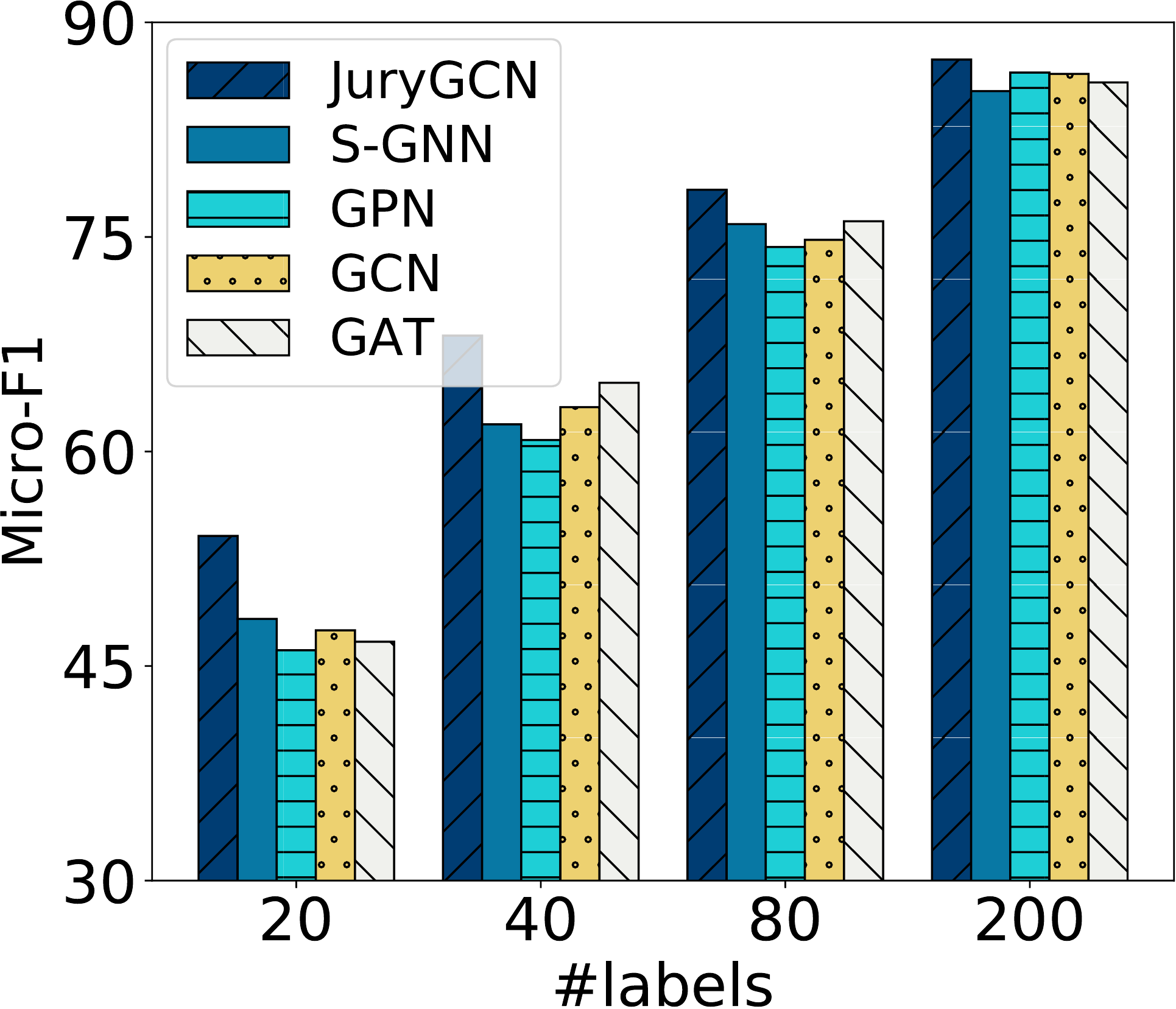}
    \vspace{-2mm}
    \caption{Reddit}
    \label{fig:results_fsl_reddit}
    \end{subfigure}}
    % \vspace{-4mm}
    \caption{Node classification results under various numbers of labels. Higher is better. Best viewed in color.}
    \vspace{-4mm}
    \label{fig:results_fsl}
\end{figure*}

\noindent\textbf{2 -- Semi-supervised Node Classification.} We classify nodes by training with various numbers of training nodes. Figure~\ref{fig:results_fsl} summarizes the results on four datasets. We can see that, in general, under the conventional setting of semi-supervised node classification (i.e., the right most bar where the number of training nodes is similar with ~\cite{kipf2017semi, velivckovic2018graph}), \method\ improves the performance of GCN to certain extent w.r.t. Micro-F1 (dark blue vs. yellow). Nonetheless, GCN can achieve a slightly better performance than \method~ on \textit{Cora}. In the mean time, as the number of training nodes becomes smaller, we can observe a consistent superiority of the proposed \method~ over other methods. For example, on \textit{Citeseer}, when $20$ training labels are provided, \method\ outperforms the best competing method (i.e., GPN) by $2.3\%$ w.r.t. Micro-F1, which further demonstrate the effectiveness of \method~ in quantifying uncertainty when the training labels are sparse. 

To summarize, the uncertainty obtained by the proposed \method\ is mostly valuable when either the total query budget (for active learning) or the total available labels (for semi-supervised node classification) is small. This is of high-importance especially for high-stake applications (e.g., medical image segmentation) where the cost of obtaining high-quality labels is high.

\vspace{-3mm}
\subsection{Efficiency Results (RQ2)}
\begin{figure}[!ht]
    \centering
    \scalebox{0.9}{
    \begin{subfigure}{.49\linewidth}
    \centering
    \includegraphics[width=.96\linewidth]{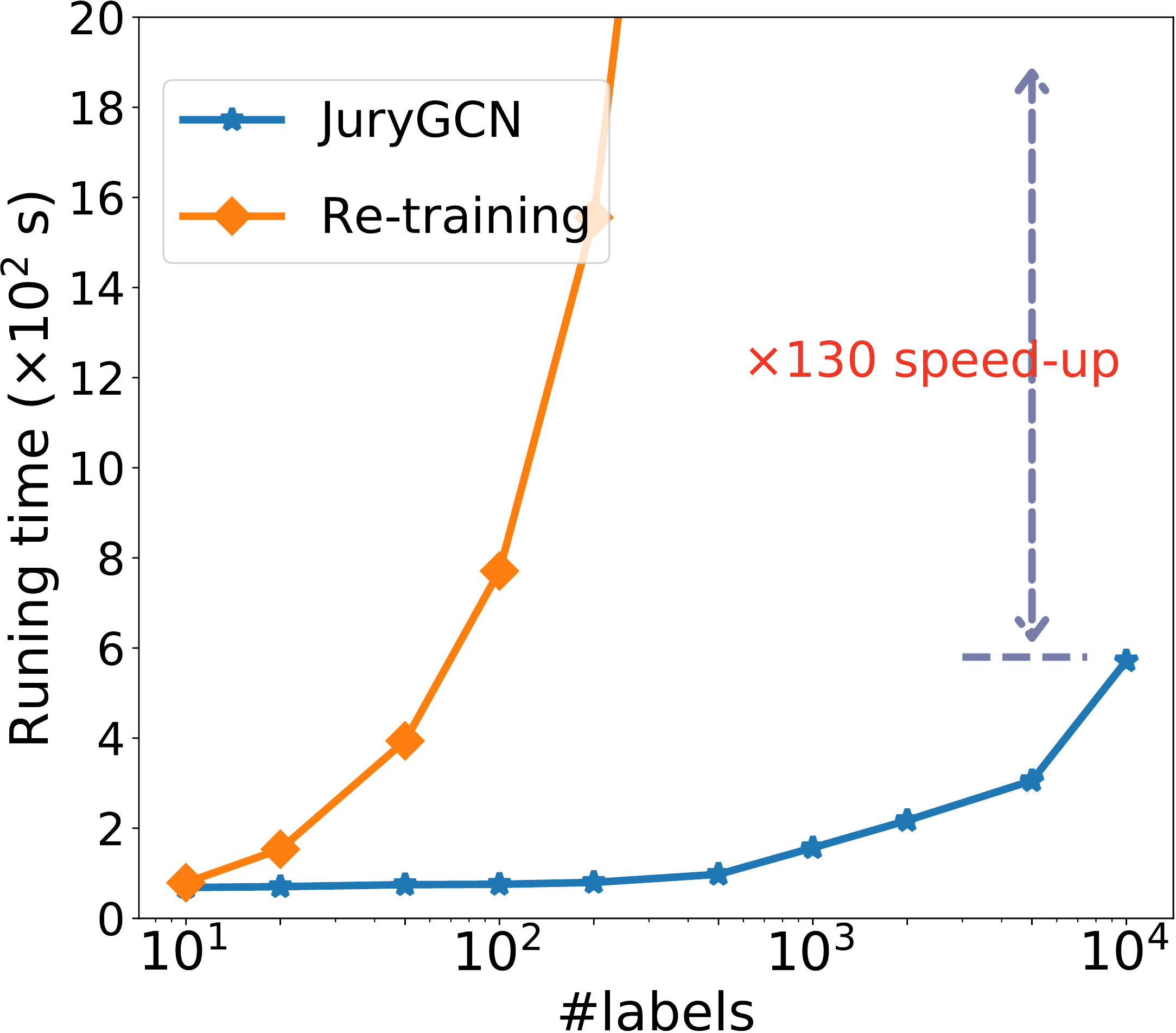}
    \vspace{-2mm}
    \caption{Running time vs. \# labels.}
    \label{fig:results_runningtime}
    \end{subfigure}%
    \begin{subfigure}{.49\linewidth}
    \centering
    \includegraphics[width=.96\linewidth]{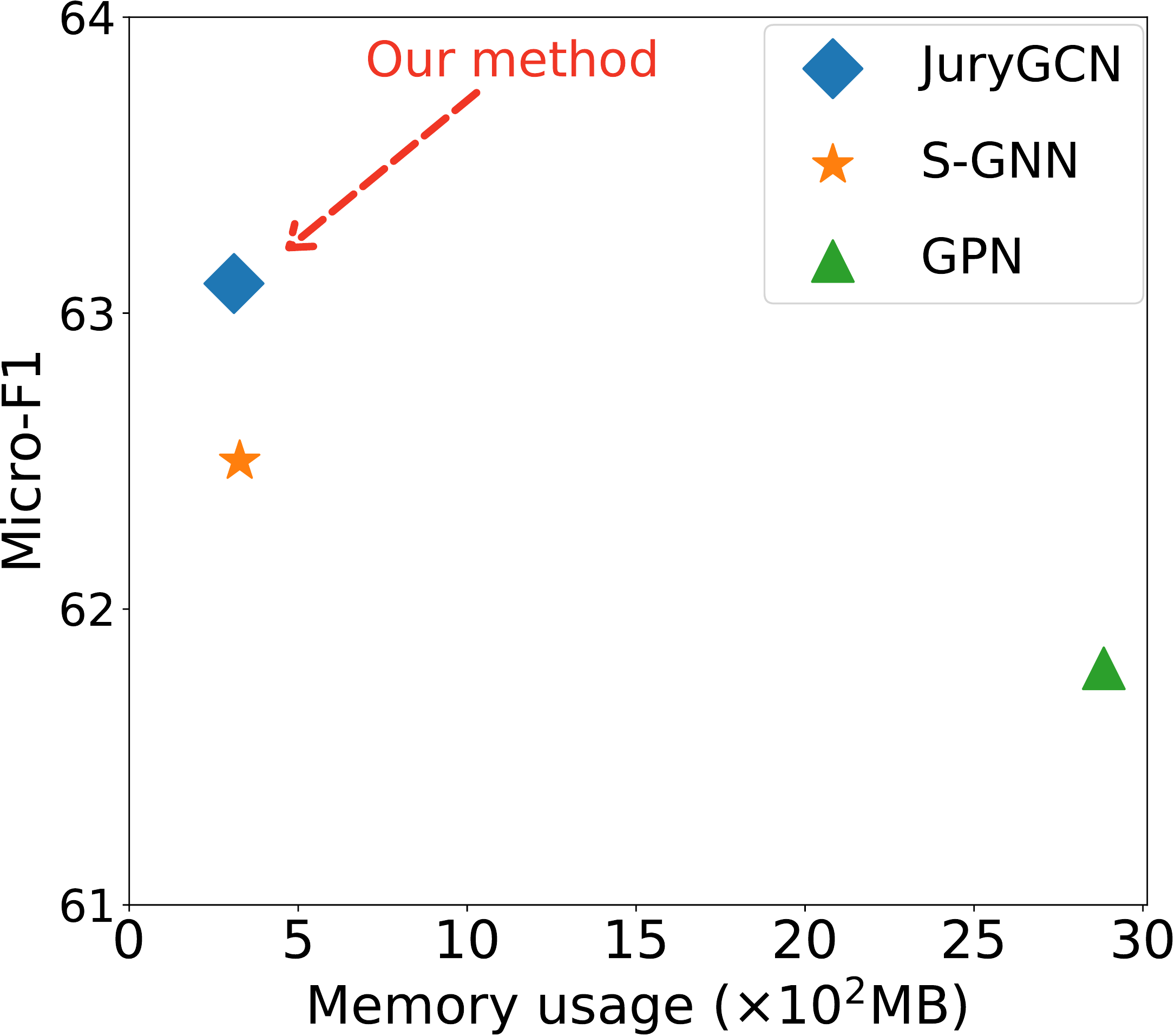}
    \vspace{-2mm}
    \caption{Micro-F1 vs. memory usage.\hide{\jian{add a caption?}}}
    \label{fig:results_acc_vs_memory}
    \end{subfigure}}
    % \vspace{-4mm}
    \caption{Efficiency results w.r.t. time and memory usage.}
    % \vspace{-4mm}
    \label{fig:results_efficiency}
\end{figure}

We evaluate the efficiency of \method\ on \textit{Reddit} in terms of running time and memory usage (Figure~\ref{fig:results_efficiency}). Regarding running time, we compare the influence function-based estimation with re-training when leaving out one sample at a time. In Figure~\ref{fig:results_runningtime}, as the number of training nodes increases, the total running time of re-training becomes significantly larger than that of \method. For example, \method~ can achieve over $130\times$ speed-up over re-training with $10,000$ training labels. In terms of memory usage in Figure~\ref{fig:results_acc_vs_memory}, compared to GPN and S-GNN, \method~(i.e., blue diamond at the upper-left corner) reaches the best balance between Micro-F1 and memory usage, with the best effectiveness and low memory usage.
% Other than influence function-based estimation, a straightforward method for uncertainty quantification is re-training the model after leaving out one sample at a time. In this section, we present the efficiency performance by comparing (1) the running time between \method~ and re-training for quantifying the GCN uncertainty on \textit{Reddit} dataset, and (2) the balance between effectiveness and process memory usage when running the experiments on \textit{Citeseer} datasets. The corresponding results are summarized in Figure~\ref{fig:results_efficiency}. Specifically, we can clearly find that in Figure~\ref{fig:results_runningtime}, as the number of training nodes increases, the total running time of re-training becomes significantly larger than that of \method. For example, \method~ can achieve over $130\times$ speed-up over re-training when the number of labels is $10,000$. In terms of memory usage, compared to GPN and S-GNN, we can see that \method~(i.e., blue diamond at the upper-left corner) reaches a satisfactory balance between classification accuracy and memory usage, with the best effectiveness performance and low memory usage.

\vspace{-3mm}
\subsection{Parameter and Sensitivity Analysis (RQ3)}
\begin{figure}[!ht]
    \centering
    \scalebox{0.9}{
    \begin{subfigure}{.49\linewidth}
    \centering
    \includegraphics[width=.96\linewidth]{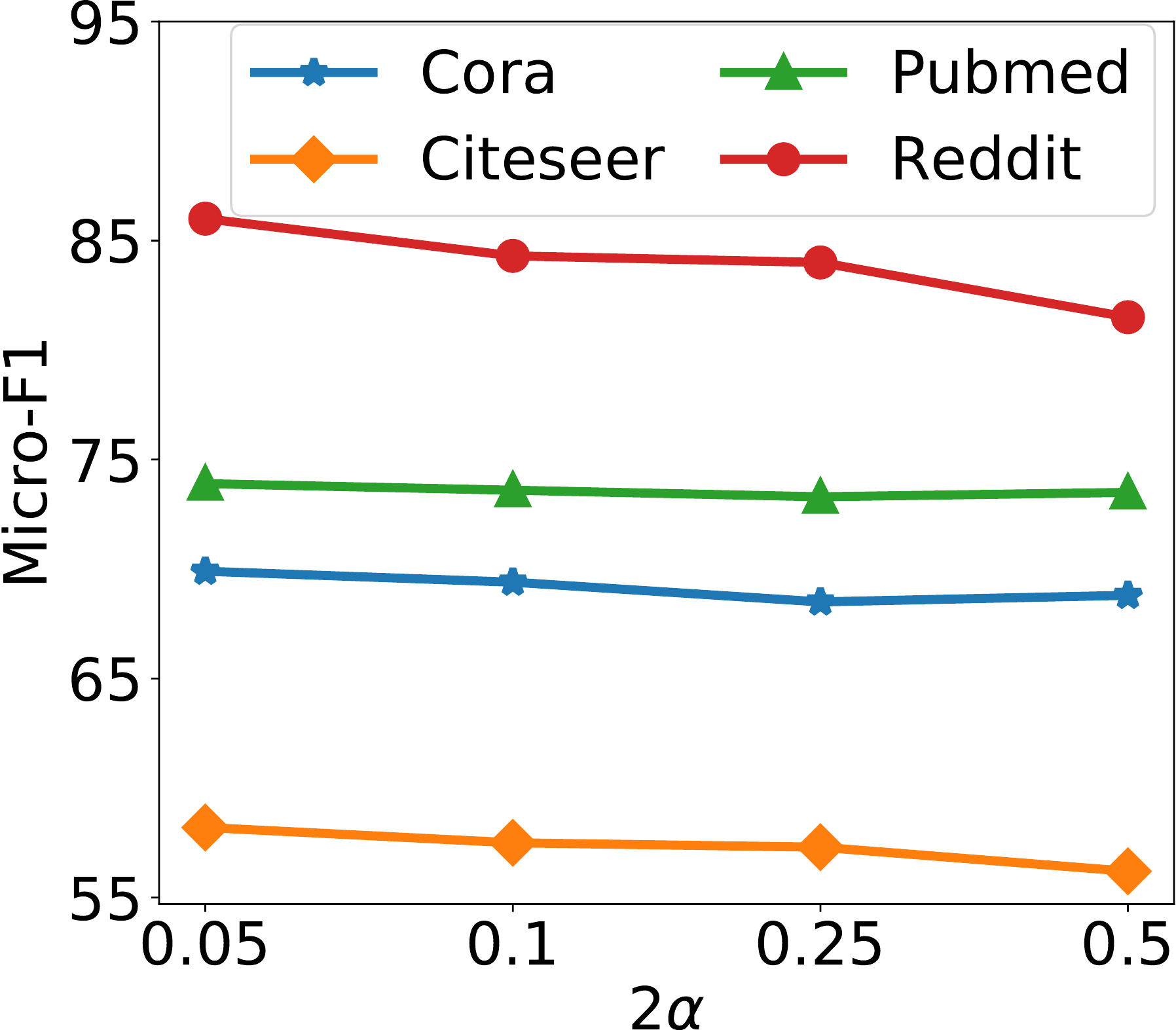}
    \vspace{-2mm}
    \caption{Micro-F1 vs. $2\alpha$. }
    \label{fig:results_para_coverage}
    \end{subfigure}%
    \begin{subfigure}{.49\linewidth}
    \centering
    \includegraphics[width=.96\linewidth]{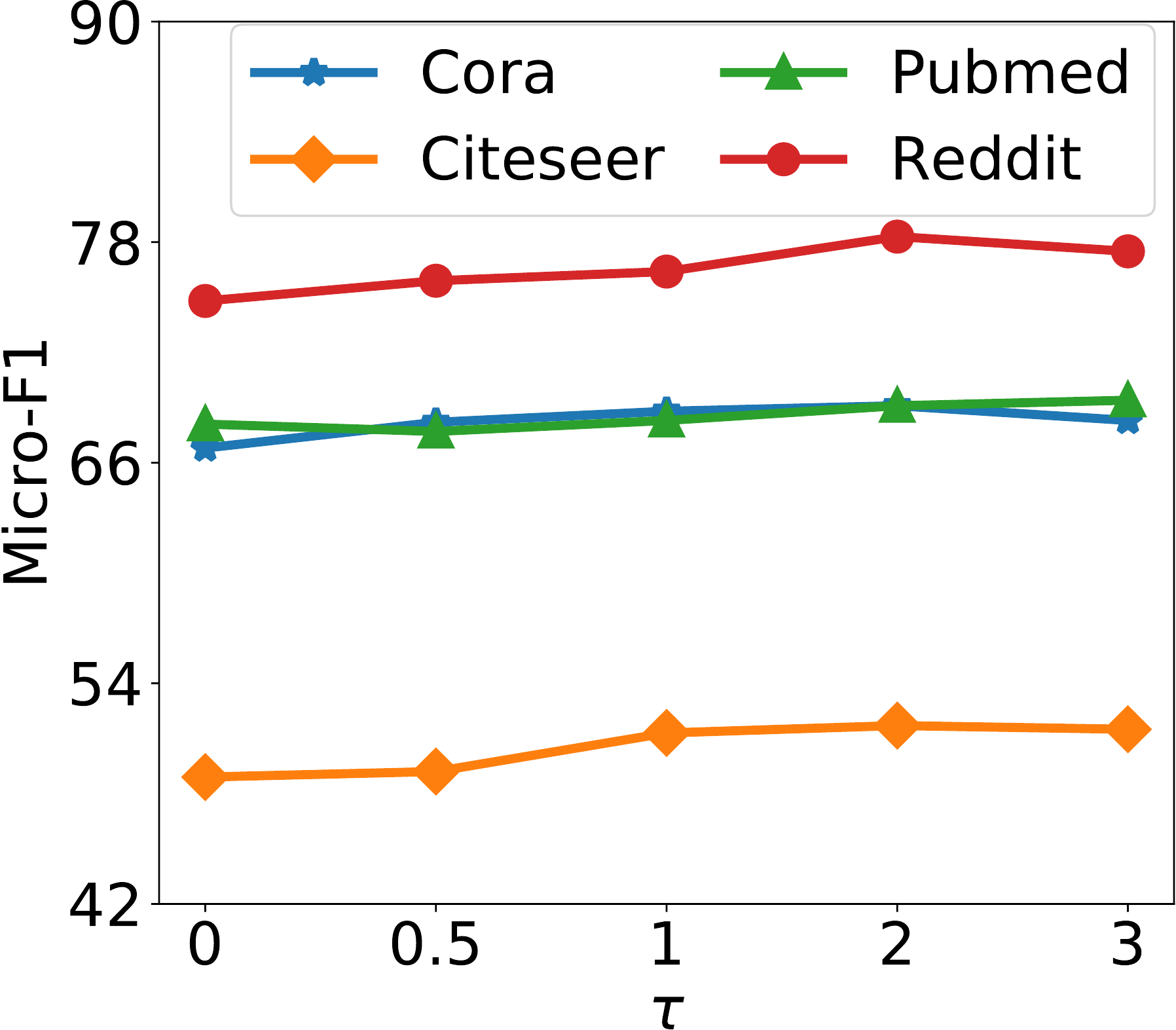}
    \vspace{-2mm}
    \caption{Micro-F1 vs. $\tau$.}
    \label{fig:results_para_tau}
    \end{subfigure}}
    % \vspace{-4mm}
    \caption{Parameter study on the coverage parameter $\alpha$ and the hyperparameter $\tau$.}
    % \vspace{-6mm}
    \label{fig:results_parameter}
\end{figure}

We investigate the sensitivity of \method\ w.r.t. (1) the coverage parameter $\alpha$ in active learning at the third query step and (2) the hyperparameter $\tau$ %(Eq.~\eqref{eq:uncertainty-based_loss}) 
in semi-supervised node classification, where the number of training labels corresponds to the third value on x-axis in Figure~\ref{fig:results_fsl}. Figure~\ref{fig:results_parameter} presents the results of sensitivity analysis. For active learning on node classification (i.e., Figure~\ref{fig:results_para_coverage}), the Micro-F1 results represent the performance at the third query step (i.e., the middle line of each dataset in Table~\ref{tab:results_al}). And for semi-supervised classification, the number of labels corresponds to the third value on x-axis in Figure~\ref{fig:results_fsl}. Regarding the sensitivity of $\alpha$, results in Figure~\ref{fig:results_para_coverage} shows that Micro-F1 slightly decreases as $\alpha$ increases. It might because smaller $\alpha$ indicates a larger target coverage ($1-2\alpha$)\footnote{The theoretical coverage of jackknife+ is $1-2\alpha$.}, resulting in wider confidence intervals. Hence, different levels of uncertainty can be accurately captured for selecting valuable query nodes. Regarding the sensitivity of $\tau$% in Eq.~\eqref{eq:uncertainty-based_loss}
, we can observe from Figure~\ref{fig:results_para_tau} that \method\ is comparatively robust to $\tau$ from $0$ (i.e., cross-entropy loss) to $3$. Meanwhile, by adjusting the importance of training node with the scale factor, Micro-F1 of \method\ improves, which is consistent with our findings in Section~\ref{subsec:effectiveness_results}.

\vspace{-2mm}
\section{Related Work}~\label{sec:related}
% In this section, we briefly review related literature for (1) graph neural networks and (2) uncertainty quantification.
\noindent\textbf{Graph neural networks} (GNNs) often reveal the state-of-the-art empirical performance on many tasks like classification~\cite{kipf2017semi}, anomaly detection~\cite{ding2021few} and recommendation~\cite{zhou2019towards,wang2019neural}. 
Bruna et al.~\cite{bruna2014spectral} leverages a learnable diagonal matrix to simulate the convolution operation in graph signal processing. 
Defferrard et al.~\cite{defferrard2016convolutional} improve the efficieny of convolution operation on graphs with the Chebyshev expansion of the graph Laplacian.
Kipf and Welling~\cite{kipf2017semi} approximates the spectral graph convolution with neighborhood aggregation over one-hop neighbors. 
Hamilton et al.~\cite{hamilton2017inductive} inductively learn node representations by sampling and aggregating node representations over the local neighborhood of a node.
Veli{\v{c}}kovi{\'c} et al.~\cite{velivckovic2018graph} introduce the self-attention mechanism to graph neural networks.
Chen et al.~\cite{chen2018fastgcn} enable batch training on GCN by sampling the receptive fields in each hidden layer.
Rong et al.~\cite{rong2019dropedge} drops a certain number of edges during each training epoch of a graph neural network.
Different from \cite{chen2018fastgcn, rong2019dropedge} that utlizes dropout or sampling based strategies to learn node representations with uncertainty, our work deterministically quantifies the uncertainty in a post-hoc manner.
Wang et al.~\cite{wang2021confident} calibrate the confidence of GCN by utilizing another GCN as the calibration function. 
Our work differs from \cite{wang2021confident} that we do not rely on any additional model to learn the confidence interval of GCN. 
We refer to recent survey~\cite{zhou2020graph} for more related works on GNNs.

\noindent\textbf{Uncertainty quantification} aims to understand to what extent a model is likely to misclassify a data sample. There has been a rich collection of research works in quantifying uncertainty for IID data~\cite{gal2016dropout, lakshminarayanan2017simple, malinin2018predictive, maddox2019simple, van2020uncertainty, alaa2020discriminative}. More related works on IID data can be found in recent survey~\cite{abdar2021review}.
Regarding uncertainty quantification for graph data, 
Dallachiesa et al.~\cite{dallachiesa2014node} classify nodes in consideration of uncertainty in edge existence.
Hu et al.~\cite{hu2017embedding} learns node embeddings for an undirected uncertain graph.
Eswaran et al.~\cite{eswaran2017power} use Dirichlet distribution to model uncertainty in belief propagation.
Ng et al.~\cite{ng2018bayesian} propose Graph Gaussian Process (GGP) as a Bayesian linear model on the feature maps of nodes.
Liu et al.~\cite{liu2020uncertainty} further extend GGP by considering input graph as an uncertain graph.
Zhang et al.~\cite{zhang2019bayesian} propose Bayesian Graph Convolutional Neural Networks which infer the joint posterior of node labels given the random graph and weight parameters.
Hasanzadeh et al.~\cite{hasanzadeh2020bayesian} train graph neural networks with random mask for each edge in each layer drawn from a Bernoulli distribution.
Different from \cite{dallachiesa2014node, hu2017embedding, eswaran2017power, zhang2019bayesian, hasanzadeh2020bayesian}, our work deterministically quantifies the uncertainty using influence functions.
Zhao et al.~\cite{zhao2020uncertainty} propose Subjective GNN (S-GNN) to model uncertainty of graph neural networks in both deep learning and subjective logic domain. 
Stadler et al.~\cite{stadler2021graph} propose Graph Posterior Netwokr, which extends the Posterior Network~\cite{charpentier2020posterior} to graphs for uncertainty estimation.
Compared with \cite{zhao2020uncertainty, stadler2021graph}, our work quantifies uncertainty in a post-hoc manner without changing the training procedures of GCN.%\hh{crazy thinking and we probably do not even need to mention this here, but does the work on box embedding (e.g., query2box, Lihui's newlook, that generates a box instead of a vector/point for each node) or geometric embedding (that generates a distribution instead of a vector/point for each node) provide a way to quantify uncertainty?}\jian{i haven't dig into details about works on box embedding, but will keep it in my note and check later.}

\vspace{-3mm}
\section{Conclusion}~\label{sec:conclusion}
In this paper, we study the problem of jackknife uncertainty quantification on Graph Convolutional Network (GCN) from the frequentist perspective. We formally define the jackknife uncertainty of a node as the width of confidence interval by a jackknife (leave-one-out) estimator. To scale up the computation, we rely on the influence functions for efficient estimation of the leave-one-out parameters without re-training. The proposed \method\ framework is 
applied to both active learning, where the most uncertain nodes are selected to query the oracle, and semi-supervised node classification, where the jackknife uncertainty serves as the importance of loss to focus on nodes with high uncertainty. Extensive evaluations on real-world datasets demonstrate the efficacy of \method\ in both active learning and semi-supervised node classification. Our proposed \method\ is able to generalize on other learning tasks beyond GCN, which is the future direction we would like to investigate.
\vspace{-3mm}
\section*{Acknowledgement}
This work is supported by National Science Foundation under grant No. 1947135, %hh-career-new
and 2134079 %MoDL 
by the NSF Program on Fairness in AI in collaboration with Amazon under award No. 1939725, % FAI
by DARPA HR001121C0165, %DARPA INCAS
% by Agriculture and Food Research Initiative (AFRI) grant No. 2020-67021-32799/project accession No.1024178 from the USDA National Institute of Food and Agriculture, %AIFarm 
% shorter version: 
by NIFA award 2020-67021-32799, % NSF-NIFA AIFarm@UIUC
and Army Research Office (W911NF2110088). %durip
The content of the information in this document does not necessarily reflect the position or the policy of the Government or Amazon, and no official endorsement should be inferred. The U.S. Government is authorized to reproduce and distribute reprints for Government purposes notwithstanding any copyright notation here on.

\vspace{-3mm}
\bibliographystyle{ACM-Reference-Format}
\bibliography{ref}

\eject
\clearpage
\section*{Appendix}
\subsection*{Additional Experimental Settings}
\noindent \textbf{1 -- Comparison Methods.} In this section, we present the detailed description of comparison methods that are used in the experiments. For the application of active learning on node classification, we have the following methods.

\begin{itemize}[
    align=left,
    leftmargin=2em,
    itemindent=0pt,
    labelsep=0pt,
    labelwidth=1em,
]
    \item \textbf{AGE}~\cite{cai2017active} measures the informativeness of each node by considering the following perspectives, including (1) the entropy of the prediction results, (2) the centrality score, and (3) the distance between the corresponding representation and its nearest cluster center. At each step of query, nodes with the highest scores are selected.
    \item \textbf{ANRMAB}~\cite{gao2018active} leverages the same selection criterion as in AGE. Additionally, ANRMAB proposes a multi-armed bandit framework to dynamically adjust weights for the three perspectives. ANRMAB utilizes the performance score of previous query steps as the rewards to learn the optimal combination of weights during the query process.
    \item \textbf{Coreset}~\cite{sener2017active} is originally proposed for Convolutional Neural networks and performs k-means clustering on the vector representations from the last hidden layer. We follow Hu et al.~\cite{hu2020graph} and apply Coreset on the node representations obtained by GCN. At each query step, we select the node which is the closet to the cluster center to label.
    \item \textbf{Centrality} selects the nodes with the highest scores of betweenness centrality at each query step. %At each query step, we select the nodes with the highest scores of betweenness centrality  for annotation.
    \item \textbf{Degree} queries the node with the highest degree. %selects the node with the highest degree for query at each step. %At each step, node with the largest degree is selected for query.
    \item \textbf{Random} annotates node randomly at each query step.
    \item \textbf{SOPT-GCN}~\cite{ng2018bayesian} utilizes $\Sigma$-optimal (SOPT) acquisition function as the active learner~\cite{ma2013sigma}, which requires the graph Laplacian and the indices of labeled nodes. We follow the same setting as in ~\cite{ng2018bayesian} to conduct the experiments.
\end{itemize}

For semi-supervised node classification, we compare the proposed framework with the following approaches.
% \jian{might need some transition word, methods below are for node classification}
\begin{itemize}
    \item \textbf{S-GNN}~\cite{zhao2020uncertainty} is an uncertainty-aware estimation framework which leverages a graph-based kernel Dirichlet distribution to estimate different types of uncertainty associated with the prediction results. We utilize the obtained node-level representations to perform classification in the experiments.
    \item \textbf{GPN}~\cite{stadler2021graph} derives three axioms for characterizing the predictive uncertainty and proposes Graph Posterior Network (GPN) to perform Bayesian posterior updates over predictions based on density estimation and diffusion. Similarly, we utilize the node representations for classification task.
    \item \textbf{GCN}~\cite{kipf2017semi} learns node-level representations by stacking multiple layers of spectral graph convolution.
    \item \textbf{GAT}~\cite{velivckovic2018graph} computes the representation of each node by introducing the learnable attention weights from its neighbors.
\end{itemize}

% \subsection*{Implementation Details}\label{sec:appendix_implementation}
\noindent \textbf{2 -- Implementation Details.} In the experiment, we conduct empirical evaluations in two applications, including (1) Application \#1: active learning on node classification, and (2) Application \#2: semi-supervised node classification, as described in Section~\ref{subsec:application}. 

In Application \#1, we evaluate all methods on a randomly constructed test set of  $1,000$ nodes for \textit{Cora}, \textit{Citeseer}, \textit{Pubmed} datasets and $139,779$ nodes for \textit{Reddit}. The validation sets for the first three citation networks contains $500$ nodes, and $23,296$ is the size of validation set for \textit{Reddit}. The remaining nodes comprise the training set (i.e., $\mathcal{V}_{\mathrm{train}}$) where the nodes for query are selected. For \textit{Cora}, \textit{Citeseer}, \textit{Pubmed} and \textit{Reddit}, (1) we have $10$, $10$, $5$ and $20$ randomly selected labels, respectively, to initiate the computation of jackknife uncertainty using Algorithm~\ref{alg:uq}; (2) the query budgets are set as $100$, $100$, $50$ and $250$; and (3) the query step sizes are $20$, $20$, $10$ and $50$.

In Application \#2, we randomly selected $100$, $100$, $50$ and $200$ nodes from \textit{Cora}, \textit{Citeseer}, \textit{Pubmed} and \textit{Reddit} respectively as the training nodes. The sizes of test sets are the same as those in Application 1. During training, we run Algorithm~\ref{alg:uq} to perform uncertainty estimation over the training nodes every 10 epochs and update the scale factor $\alpha$ accordingly. In addition, we also evaluate the model performance when the number of training nodes is significantly small.

In both applications, we adopt a two-layer GCN with 16 hidden layer dimension. For training, we use the Adam optimizer~\cite{kingma2017adam} with learning rate $0.01$ and train the GCN classifier for 100 epochs. The coverage parameter ($\alpha$) in Algorithm~\ref{alg:uq} is $0.025$ and the hyperparameter $\tau$ %in Eq.~\eqref{eq:uncertainty-based_loss} 
in Application 2 is set as $2$. For all other comparison methods, we use the original settings \hide{to perform evaluation}. We report the average results after 20 runs of each method on two applications.

\section*{Proof of Proposition~\ref{prop:both_types_on_GCN}}
For an arbitrary node $v\in\mathcal{V}$ in the graph, its receptive field after $L$ graph convolution layers is the set of all neighbors within $L$ hops. Then if the node $u$, whose uncertainty is going to be quantified, is not within the $L$-hop neighborhood of any training node $v\in\mathcal{V}_{\textrm{train}}\setminus\{u\}$, whether to leave out the loss of node $u$ during training will have no impact on hidden node representations and final predictions of $v$, which is equivalent to the leave-one-out settings for IID data.

\section*{Proof of Proposition~\ref{lm:first_order}}
To prove it, we derive the element-wise computation of $\nabla_{\mathbf{W}^{(l)}} r(i, \mathbf{y}_i, \Theta)$ and then write out the matrix form. We first apply the chain rule and get
\begin{equation}\label{eq:first_order_grad}
	\frac{\partial r(i, \mathbf{y}_i, \Theta)}{\partial \mathbf{W}^{(l)}[a, b]} = \sum_{c=1}^n \sum_{d=1}^{h_l} \frac{\partial r(i, \mathbf{y}_i, \Theta)}{\partial \mathbf{E}^{(l)} [c, d]} \frac{\partial \mathbf{E}^{(l)} [c, d]}{\partial \mathbf{W}^{(l)}[a, b]}
\end{equation}
where $h_l$ is the hidden dimension of $l$-th layer. Regarding the computation of $\frac{\partial \mathbf{E}^{(l)} [c, d]}{\partial \mathbf{W}^{(l)}[a, b]}$, since $\mathbf{E}^{(l)} = \sigma(\mathbf{\hat A} \mathbf{E}^{(l-1)} \mathbf{W}^{(l)})$, we have
\begin{equation}\label{eq:key_term_2}
\begin{aligned}
	\frac{\partial \mathbf{E}^{(l)} [c, d]}{\partial \mathbf{W}^{(l)}[a, b]} 
	& = \frac{\partial \sigma\big(\mathbf{\hat A} \mathbf{E}^{(l-1)} \mathbf{W}^{(l)}\big)[c, d]}{\partial \big(\mathbf{\hat A} \mathbf{E}^{(l-1)} \mathbf{W}^{(l)}\big)[c, d]} 
		\frac{\partial \big(\mathbf{\hat A} \mathbf{E}^{(l-1)} \mathbf{W}^{(l)}\big)[c, d]}{\partial \mathbf{W}^{(l)}[a, b]} \\
	& = \sigma'\big(\mathbf{\hat A} \mathbf{E}^{(l-1)} \mathbf{W}^{(l)}\big)[c, d]
		\big(\mathbf{\hat A} \mathbf{E}^{(l-1)}\big)[c, a] \mathbf{I}[d, b]
\end{aligned}
\end{equation}
where $\sigma '$ is the first-order derivative of the activation function $\sigma$. Combining Eqs.~\eqref{eq:first_order_grad} and ~\eqref{eq:key_term_2} together, we have the element-wise computation of $\nabla_{\mathbf{W}^{(l)}} r(i, \mathbf{y}_i, \Theta)$ as follows.
\begin{equation}\label{eq:first_order_elementwise}
	\frac{\partial r(i, \mathbf{y}_i, \Theta)}{\partial \mathbf{W}^{(l)}[a, b]} 
	= \big(\mathbf{\hat A}\mathbf{E}^{(l-1)}\big)^T[a, :] 
		\bigg(\frac{\partial r(i, \mathbf{y}_i, \Theta)}{\partial \mathbf{E}^{(l)}} \circ  
		\sigma'\big(\mathbf{\hat A} \mathbf{E}^{(l-1)} \mathbf{W}^{(l)}\big)\bigg)[:, b]
\end{equation}
where $\circ$ is the element-wise product. Finally, we get Eq.~\eqref{eq:first_order} by writing Eq.~\eqref{eq:first_order_elementwise} into matrix form.

Regarding the computation of $\frac{\partial r(i, \mathbf{y}_i, \Theta)}{\partial \mathbf{E}^{(l)}}$, the key idea is to write out a recursive function with respect to $\frac{\partial r(i, \mathbf{y}_i, \Theta)}{\partial \mathbf{E}^{(l)}}$ based on the chain rule. More specifically, we first consider the element-wise computation and apply the chain rule as follows.
\begin{equation}\label{eq:derivative_wrt_embeddings}
    \frac{\partial r(i, \mathbf{y}_i, \Theta)}{\partial \mathbf{E}^{(l)} [c, d]} = 
    \sum_{e=1}^{n}\sum_{f=1}^{h_{l+1}} 
        \frac{\partial r(i, \mathbf{y}_i, \Theta)}{\partial \mathbf{E}^{(l+1)} [e, f]}
        \frac{\partial \mathbf{E}^{(l+1)} [e, f]}{\partial \mathbf{E}^{(l)} [c, d]}
\end{equation}
We take derivative on both sides of $\mathbf{E}^{(l)} = \sigma(\mathbf{\hat A} \mathbf{E}^{(l-1)} \mathbf{W}^{(l)})$ and get
\begin{equation}\label{eq:derivative_between_embeddings}
    \frac{\partial \mathbf{E}^{(l+1)} [e, f]}{\partial \mathbf{E}^{(l)} [c, d]} = 
        \sigma'\big(\mathbf{\hat A} \mathbf{E}^{(l)} \mathbf{W}^{(l+1)}\big)[e,f]
        \mathbf{\hat A}[e, c]
        \mathbf{W}^{(l+1)}[d, f]
\end{equation}
Combining Eqs.~\eqref{eq:derivative_wrt_embeddings} and \eqref{eq:derivative_between_embeddings} together, we get the following element-wise first-order derivative.
\begin{equation}\label{eq:first_order_embedding_elementwise}
\begin{aligned}
    \frac{\partial r(i, \mathbf{y}_i, \Theta)}{\partial \mathbf{E}^{(l)} [c, d]} = \sum_{e=1}^{n}\sum_{f=1}^{h_{l+1}} 
        & \mathbf{\hat A}^T[c, e] \cdot
        \bigg(
            \frac{\partial r(i, \mathbf{y}_i, \Theta)}{\partial \mathbf{E}^{(l+1)}}
            \sigma'\big(\mathbf{\hat A} \mathbf{E}^{(l)} \mathbf{W}^{(l+1)}\big)
        \bigg) [e, f] \\
        & \cdot \big(\mathbf{W}^{(l+1)}\big)^T[f, d]
\end{aligned}
\end{equation}
Written Eq.~\eqref{eq:first_order_embedding_elementwise} into matrix form, we complete the proof.

\section*{Proof of Theorem~\ref{lm:second_order}}
We prove case by case.
    
    \textbf{Case 1.} When $i = l$, since the activation function $\sigma$ is the ReLU function, it is trivial that the subgradient of its second-order derivative is always 0 since the first-order derivative is the indicator function. Thus, $\mathfrak{H}_{l,l} = 0$.
    
    \textbf{Case 2.} When $i = l-1$, to get Eq.~\eqref{eq:second_order_i=l-1}, it is trivial to prove by taking derivative on both sides of Eq.~\eqref{eq:first_order}. See the proof of Proposition~\ref{lm:first_order} for the proof of Eq.~\eqref{eq:case_2_key_term}. 
    
    \textbf{Case 3.} When $i < l-1$, we first take derivative on both sides of Eq.~\eqref{eq:first_order} in $i$-th hidden layer. Then we have
    \begin{equation}\label{eq:case_3}
        \mathfrak{H}_{l,i}[:,:,c,d] = 
            \bigg(\mathbf{\hat A}\frac{\partial \mathbf{E}^{(i-1)}}{\partial \mathbf{W}^{(i)}[c,d]}\bigg)^T
            \bigg(
                \frac{\partial R}{\partial \mathbf{E}^{(i)}}
                \circ
                \sigma'_l
            \bigg)
    \end{equation}
    To compute $\frac{\partial \mathbf{E}^{(l-1)}}{\partial \mathbf{W}^{(i)}[c,d]}$, we first consider an arbitrary $(a,b)$-th element in $\frac{\partial \mathbf{E}^{(l-1)}}{\partial \mathbf{W}^{(i)}[c,d]}$, i.e., $\frac{\partial \mathbf{E}^{(l-1)}[a,b]}{\partial \mathbf{W}^{(i)}[c,d]}$. Then we take the derivative on both sides of $\frac{\partial \mathbf{E}^{(l-1)}[a,b]}{\partial \mathbf{W}^{(i)}[c,d]}$, which gives us
    \begin{equation}\label{eq:case_3_chain_rule}
    \begin{aligned}
        \frac{\partial \mathbf{E}^{(l-1)}[a,b]}{\partial \mathbf{W}^{(i)}[c,d]} & = 
        \sigma'_{l-1} [a,b]
        \cdot
        \sum_{e=1}^{n}\sum_{f=1}^{h_i}
            \mathbf{\hat A}[a, e] 
            \frac{\partial \mathbf{E}^{(l-2)} [e, f]}{\partial \mathbf{W}^{(i)}[c,d]} 
            \mathbf{W}^{(l-1)}[f, b] \\
        & =  \bigg(\sigma'_{l-1} \circ \bigg(\mathbf{\hat A}\frac{\partial \mathbf{E}^{(l-2)}}{\partial \mathbf{W}^{(i)}[c,d]} \mathbf{W}^{(l-1)}\bigg)\bigg)[a, b]
    \end{aligned}
    \end{equation}
    It is trivial to get Eq.~\eqref{eq:second_order_forwarding} by writing out the matrix form of Eq.~\eqref{eq:case_3_chain_rule}.
    
    \textbf{Case 4.} When $i = l+1$, to get Eq.~\eqref{eq:second_order_i=l+1}, it is trivial to prove by taking derivative on both sides of Eq.~\eqref{eq:first_order}. Regarding the computation of $\frac{\partial^2 R}{\partial \mathbf{E}^{(l)}[a,b] \partial \mathbf{W}^{(l+1)}[c, d]}$, by Eq.~\eqref{eq:first_order_wrt_embeddings}, we have 
    \begin{equation}
        \frac{\partial R}{\partial \mathbf{E}^{(l)}[a,b]} = \bigg(\mathbf{\hat A}^T
        \bigg(
            \frac{\partial R}{\partial \mathbf{E}^{(l+1)}}
            \circ
            \sigma'\big(\mathbf{\hat A} \mathbf{E}^{(l)} \mathbf{W}^{(l+1)}\big)
        \bigg)
        \big(\mathbf{W}^{(l+1)}\big)^T\bigg)[a,b]
    \end{equation}
    Then we take derivative on both sides and get 
    \begin{equation}
    \begin{aligned}
        \frac{\partial^2 R}{\partial \mathbf{E}^{(l)}[a,b] \partial \mathbf{W}^{(l+1)}[c, d]} 
            & = \sum_{e=1}^{n}\sum_{f=1}^{h_{l+1}} 
                \mathbf{\hat A}^T[a, e] \cdot \bigg(
                    \frac{\partial R}{\partial \mathbf{E}^{(l+1)}}
                    \sigma'_{l+1}
                \bigg) [e, f] \\
            &\qquad\qquad \cdot \frac{\partial \mathbf{W}^{(l+1)}[b, f]}{\partial \mathbf{W}^{(l+1)}[c, d]} \\
            & = \mathbf{I}[b,c] \sum_{e=1}^{n} \mathbf{\hat A}^T[a, e] \cdot \bigg(
                \frac{\partial R}{\partial \mathbf{E}^{(l+1)}}
                \sigma'_{l+1}
            \bigg)[e, d] \\
            & = \mathbf{I}[b, c] \bigg(\mathbf{\hat A}^T \bigg(\frac{\partial R}{\partial \mathbf{E}^{(l+1)}} \circ \sigma'_{l+1}\bigg)\bigg)[a, d]
    \end{aligned}
    \end{equation}

    \textbf{Case 5.} When $i > l+1$, %we first take the derivative on both sides of Eq.~\eqref{eq:first_order} and get $\mathfrak{H}_{l,i}[:,:,c,d] = \big(\mathbf{E}^{(l-1)}\big)^T \mathbf{\hat A}^T \bigg(\frac{\partial R}{\partial \mathbf{E}^{(l)}\mathbf{W}^{(i)}[c,d]} \circ \sigma'_l\bigg)$.
    % \begin{equation}\label{eq:case_5}
    %     \mathfrak{H}_{l,i}[:,:,c,d] = 
    %     \big(\mathbf{E}^{(l-1)}\big)^T 
    %     \mathbf{\hat A}^T
    %     \bigg(
    %         \frac{\partial R}{\partial \mathbf{E}^{(l)}\mathbf{W}^{(i)}[c,d]} \circ  
	   %     \sigma'_l
	   %\bigg)
    % \end{equation}
    we get Eq.~\eqref{eq:second_order_backward} by taking derivative on both sides of Eq.~\eqref{eq:first_order_wrt_embeddings}.
    
    Putting everything (Cases 1 -- 5) together, we complete the proof.

\end{document}